\documentclass[11pt]{article}

\usepackage[preprint]{neurips_2023} 
\usepackage[utf8]{inputenc}         
\usepackage[T1]{fontenc}            

\usepackage{amsmath}                
\usepackage{amsfonts}               
\usepackage{amssymb}                
\usepackage{nicefrac}               

\usepackage{booktabs}               
\usepackage{multirow}               
\usepackage{graphicx}               
\usepackage{adjustbox}              
\usepackage{subcaption}             

\usepackage{listings}                          
\usepackage[listings,breakable,skins]{tcolorbox}

\usepackage[ruled,vlined]{algorithm2e}  
\SetAlCapSkip{0.4em}

\usepackage{enumitem}               

\usepackage{microtype}              
\usepackage{xcolor}                 
\usepackage{xspace}                 
\usepackage{url}                    
\usepackage{placeins}
\usepackage{array}

\usepackage{hyperref}               
\hypersetup{colorlinks=true,linkcolor={blue!50!black},citecolor={blue!50!black},urlcolor={blue!60!black},pdftitle={ReTrace: Distilling Tool-Using Agents into Planner-Free Workflows}}

\newcommand{\name}{\mbox{ReTrace}\xspace}               
\newcommand{\dset}{\textup{\textsc{Internal}}\xspace}     

\newcommand{\tautwobench}{\texorpdfstring{\mbox{$\tau^2$-bench}}{tau2-bench}\xspace}

\definecolor{artifactframe}{HTML}{B8C4D0}
\definecolor{artifactback}{HTML}{F6F8FA}
\definecolor{artifacttitle}{HTML}{E4EAF1}

\lstdefinestyle{artifactstyle}{%
  basicstyle=\ttfamily\scriptsize,
  breaklines=true,
  breakautoindent=false,
  breakindent=1.2em,
  postbreak=\mbox{\textcolor{artifactframe}{$\hookrightarrow$}\space},
  columns=fullflexible,
  keepspaces=true,
  showstringspaces=false,
  aboveskip=0pt, belowskip=0pt,
  upquote=true,
}

\newtcblisting{artifact}[2][]{%
  listing only,
  listing options={style=artifactstyle},
  breakable, enhanced,
  colback=artifactback, colframe=artifactframe, boxrule=0.5pt, arc=2pt,
  left=4pt, right=3pt, top=3pt, bottom=3pt, boxsep=0pt,
  title={#2}, title after break={#2 \normalfont(continued)},
  fonttitle=\bfseries\footnotesize, coltitle=black,
  colbacktitle=artifacttitle,
  attach boxed title to top left={xshift=3mm, yshift*=-2pt},
  boxed title style={colframe=artifactframe, boxrule=0.4pt, arc=1pt,
                     left=3pt, right=3pt, top=1pt, bottom=1pt},
  top=8pt, #1,
}

\title{\name: Distilling Tool-Using Agents into Planner-Free Workflows}

\author{%
    Kaiyuan Hou\thanks{Work done during an internship at Amazon.}\ \thanks{Equal Contribution} \\
    Columbia University \\
    \texttt{kh3119@columbia.edu} \\
    \And
    Yang Ping Kuo\footnotemark[2]\ \thanks{Corresponding Author} \\
    Amazon \\
    \texttt{yangkuo@amazon.com} \\
    \And
    Indranil Bhattacharya\footnotemark[2] \\
    Amazon \\
    \texttt{bindrani@amazon.com} \\
    \And
    Rohith Nama\footnotemark[2] \\
    Amazon \\
    \texttt{rohinama@amazon.com} \\
    \And
    Arushi Gupta\footnotemark[2] \\
    Amazon \\
    \texttt{arushimg@amazon.com} \\
    \And
    Carlos D. Castillo \\
    Amazon \\
    \texttt{cdcast@amazon.com} \\
    \And
    Prathap Ramachandra \\
    Amazon \\
    \texttt{prathara@amazon.com} \\
    \And
    Xiaofan Jiang \\
    Columbia University \\
    \texttt{jiang@ee.columbia.edu} \\
    \And
    Francesco Carbone \\
    Amazon \\
    \texttt{carbonef@amazon.com}
}

\begin{document}

\maketitle

\begin{abstract}
Business process automation runs the same Standard Operating Procedure (SOP) again and again on new inputs. Large Language Model (LLM) agents can follow an SOP written in natural language, but they call an LLM to choose every step of every run, which adds latency and cost at scale. We present \name, which turns an SOP and a few logged agent runs into a reusable workflow in which code, not an LLM, selects the next step. The agent's reasoning, the explanation it writes with each tool call, often states why it took an action; the tool calls alone do not. \name extracts the condition behind each step, merges these conditions across runs into a compact rule document, and generates an executable workflow from it. Only this last stage depends on the platform, so rule documents can be compiled for different orchestration platforms. On $79$ procedures from our production platform, \name workflows reach $93.8\%$ completion, against $96.8\%$ for the agent, at $24\%$ lower cost per task. On \tautwobench, they reach $75.8\%$ task success (agent: $80.6\%$) and run in $1.3$\,s instead of $170$\,s; three rounds of repair on training failures lift $k{=}2$ workflows from $66.7\%$ to $81.2\%$. On eight SOP-Bench procedures, \name without repair trails few-shot generation ($80.6\%$ vs.\ $83.6\%$), but repair from labeled training failures lifts it to $87.8\%$ at $0.40$\,s per task, compared to $85.5\%$ in $46.6$\,s for the agent.
\end{abstract}

\section{Introduction}
\label{sec:intro}

Many business processes run thousands of times on different inputs: a support case is triaged, a form is checked, an order is escalated. Automating such a process has traditionally meant building a workflow by hand, wiring tools, inputs, and branches in code or in a workflow editor \citep{bpmn,vanderaalst2018rpa}. Large Language Model (LLM) agents remove most of this effort. A process owner writes a Standard Operating Procedure (SOP) in natural language, and the agent reads it and calls the available tools \citep{react,sopbench_li2025,sopbench,sopmaze}. The price is paid at run time: the agent calls an LLM to plan, route, and transform data at every step of every execution, which is slow and expensive at high volume.

\begin{figure*}[t]
\centering
\includegraphics[width=\linewidth]{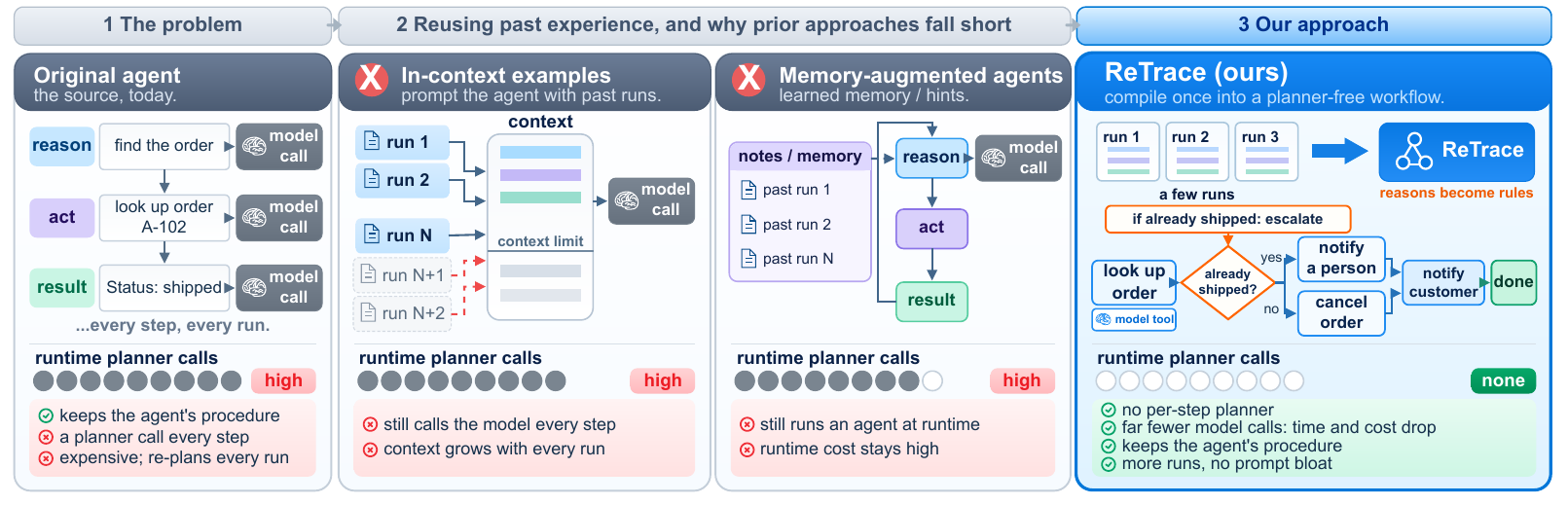}
\caption{\textbf{From agent runs to executable workflows.} The agent, and methods that reuse its past runs as examples or memory, call an LLM to choose each step. \name turns the agent's reasoning and tool calls into rules before deployment; at run time, code selects the next step.}
\label{fig:teaser}
\end{figure*}

These costs are visible in production. Of the $5{,}115$ automations that customers created on our enterprise business process automation platform between April and June 2026, $91.8\%$ were agents that plan at run time, yet seven of the ten most frequently executed automations were fixed workflows. For agents using three to five tools, choosing the next step took \mbox{$37$--$43\%$} of execution time, on top of the token cost. For a procedure that runs often, building a workflow once pays off: from the measured savings and a \tautwobench construction cost, we estimate that construction is recovered after about $75$ executions, while each of those seven workflows ran $297$k--$528$k times in three months (Appendix~\ref{app:workflow-cost}).

Logged agent runs, or \emph{traces}, show how a procedure is carried out, and prior work reuses them as written guidance, skills, or code \citep{voyager,awm,expel,skillfactory,bpop}. Replacing the agent with a workflow, however, requires knowing \emph{when} each tool should be called and \emph{where} each of its inputs comes from. Tool calls alone often do not say. If an agent looks up an order and then escalates to a human, the trace does not show whether the shipping status caused the escalation. The agent's reasoning often does, and it can also mention alternatives that no logged run took. Giving a model more raw traces is not enough: with many long traces, the workflow generator eventually writes worse code (Section~\ref{sec:results-full-traces}).

We present \name, which builds a reusable workflow from an SOP and agent traces that include reasoning (Figure~\ref{fig:teaser}). It extracts the condition, tool, and input sources behind each step, merges them across traces into a compact rule document that links every rule to its evidence, and generates an executable workflow from that document. LLMs are used only in this one-time construction; at run time, code selects the next step, though a tool may still call a model, for example to classify a document. Our contributions are:

\begin{itemize}[leftmargin=*,nosep]
\item \textbf{Rules from agent reasoning.} We combine the agent's reasoning with its tool records to recover action conditions, input sources, and alternatives that no logged run executed.

\item \textbf{One rule document, several platforms.} Only the final stage depends on the target, so the same pipeline yields Python code for two public benchmarks and typed workflow graphs for our production platform.

\item \textbf{Evidence from production and two public benchmarks.} On $79$ production procedures and on \tautwobench, \name workflows come within $3$ and $5$ points of the agent, respectively, while cutting enterprise cost per task by $24\%$ and \tautwobench time per task from $170$\,s to $1.3$\,s. Removing the agent's reasoning costs $8.8$ and $6.2$ points, respectively, and gains $1.7$ on SOP-Bench. Repaired workflows reach $81.2\%$ on \tautwobench (agent: $80.6\%$, n.s.) and $87.8\%$ on SOP-Bench (agent: $85.5\%$), where unrepaired \name trails few-shot.
\end{itemize}

\section{Related Work}
\label{sec:related}

\paragraph{Reusing agent experience.} Agent Workflow Memory (AWM) stores workflow descriptions as written guidance for an agent, while ExpeL and ReasoningBank store lessons and strategies \citep{awm,expel,reasoningbank}. Voyager, ASI, SKILL-DISCO, and NSI instead produce reusable code, or skills, that an LLM agent selects and calls at run time \citep{voyager,asi,skilldisco,nsi}. Skill Factory builds skills from successful runs, but a model still picks a skill and fills its inputs for each task \citep{skillfactory}; BPOP learns only which actions must precede others \citep{bpop}. In all of these but BPOP, a model keeps choosing steps; \name compiles conditions and input sources into a workflow that does not need one. SOP benchmarks test whether an agent follows a written procedure at run time \citep{sopbench_li2025,sopbench,sopmaze}; \name instead compiles the procedure once, so no agent has to follow it on each run.

\paragraph{Compiling executable workflows.} TraceCompiler tracks how tool outputs feed later inputs and builds mostly deterministic workflows that may still call an LLM to fill some inputs; on AppWorld, it escalates when a required branch never appeared in the logged runs \citep{tracecompiler}. Trace2Policy reconstructs experts' decision rules from behavior records with a Vision-Language Model (VLM) and revises them on validation errors; its deployed version is human-refined Python that runs without an LLM \citep{trace2policy}. Concurrent work on \tautwobench keeps a model in the loop at run time: PLVR guides or checks a small model that still generates actions, and STAGE runs model-backed judgment nodes inside a code-controlled graph \citep{plvr,stage}. \name differs in its evidence: it reads the conditions and alternatives stated in the agent's own reasoning, which can cover branches that no logged run took. Workflow search and optimization methods tune agent designs, prompts, or LLM-calling steps \citep{aflow,adas,gptswarm,lats,dspy,textgrad,trace}; we instead recover workflow rules from an SOP and logged runs. Table~\ref{tab:positioning} in Appendix~\ref{app:positioning} compares the closest methods.

\section{Method}
\label{sec:method}

\subsection{Inputs and target workflow}
\label{sec:method-setup}

A \emph{task family} is a set of inputs handled by the same SOP $P$, for example support cases under one troubleshooting procedure. We use $k$ traces $\mathcal{T}_k=\{\tau^{(1)},\dots,\tau^{(k)}\}$ of an agent on such inputs. A trace $\tau=(s_1,\dots,s_L)$ records each of its $L$ steps as $s_t=(r_t,a_t,v_t,o_t)$, where $r_t$ is the agent's reasoning for the step, $a_t$ the tool called, $v_t$ its inputs, and $o_t$ the result.

The output is an \emph{executable workflow} $\pi$, built once per procedure and reused on new inputs. At run time, code chooses the next step without calling an LLM; this is what we mean by \emph{planner-free}. Individual tools may still call models. The workflow is Python code on the public benchmarks and a typed graph of steps on \dset, our enterprise dataset.

\subsection{Building the workflow}
\label{sec:method-retrace}

Construction has three stages: \emph{extraction} reads each step, \emph{aggregation} merges evidence across traces into an \emph{Intermediate Representation} (IR) $\mathcal{I}$ (a compact rule document); \emph{synthesis} generates the workflow from it. Algorithm~\ref{alg:retrace-detail} in Appendix~\ref{app:method-construction} gives the full procedure.

\paragraph{Extraction: reading each step.}
\label{sec:method-extract}

For each step, an LLM extracts a record $z_t=(g_t,\alpha_t,a_t,v_t,\omega_t)$. The reasoning supplies the condition $g_t$ under which the action applies and the intended action $\alpha_t$. The condition may use only information available before the action: ``airplane mode is on'' can justify turning it off, but the result of turning it off cannot. The tool record supplies the tool $a_t$, its typed inputs $v_t$, and the source $\omega_t$ of each input, such as a task input, an earlier result, or a constant; unknown sources are marked as such. Case-specific values, such as a customer ID, become variables filled from each new task. The record also keeps alternative actions that the reasoning mentions but the agent did not take.

\paragraph{Aggregation: combining traces into rules.}
\label{sec:method-aggregate}

An LLM first summarizes each trace against the SOP, matching its steps to the procedure and noting alternatives that were mentioned but not executed. A merge step then combines the summaries: it unifies repeated steps, works out how earlier results feed later inputs, checks the reasoning against the recorded calls and outcomes, and resolves disagreements using the SOP, giving a reason for anything it drops. The result is a completion check and an ordered list of rules:
\begin{equation}
  \mathcal{I}=\big(\chi,\langle(g_1,\alpha_1),\dots,(g_m,\alpha_m)\rangle\big).
  \label{eq:ir}
\end{equation}
Here $\chi$ tests whether the task is complete, and rule $j$ pairs a condition, or \emph{guard}, $g_j$ with an action $\alpha_j$, its tool, and its input sources. The rules form a decision list \citep{decisionlists}. Each rule cites the trace steps or SOP statements that support it, and alternatives that no trace executed are marked \emph{unwitnessed}; for example, the reasoning for turning off airplane mode may also name disabled mobile data as a cause, which becomes an unwitnessed rule (Appendix~\ref{app:authorinput}). A final check uses rule names and citations to confirm that no rule or named alternative was dropped silently; it does not judge whether a condition is right. Beyond eight traces, the rest are added one at a time (Appendix~\ref{app:aggregation-details}). The workflow generator reads the IR, not the raw traces; on \tautwobench at $k{=}16$, its input is $24.5$k tokens, against $101.4$k with full traces (Table~\ref{tab:carrier} in Appendix~\ref{app:authorinput}).

\paragraph{Synthesis: generating and checking the workflow.}
\label{sec:method-synthesize}

An LLM converts the IR into code or the platform's workflow format. Let $h$ be the state of a run: the task input plus every tool call and result so far. At each step, the workflow stops if the completion check succeeds; otherwise, it runs the first rule whose guard holds:
\begin{equation}
  \pi(h)=
  \begin{cases}
    \texttt{stop} & \text{if }\ \chi(h),\\[2pt]
    \alpha_{j^\star(h)}(h) & \text{otherwise},
  \end{cases}
  \label{eq:program}
\end{equation}
where $j^\star(h)=\min\{j:g_j(h)\text{ holds}\}$. The action calls its tool with inputs taken from $h$ or constants, and the result updates $h$. If the completion check fails and no rule applies, the workflow stops as incomplete and never asks an LLM for a step.

Before deployment, generated code must load and must not be able to make planning calls. On the public benchmarks, candidates may import only approved libraries and run in a sandbox; we reject any candidate that makes a planning call when re-executed on the training inputs and keep the one that reproduces the most recorded steps. On \dset, the platform builds a single workflow graph, which must contain no node that asks a model for the next step. Neither check establishes task correctness (Appendix~\ref{app:guarantees}). Where the benchmark can be re-executed offline, the workflow can also be repaired from its failures on training inputs: an LLM revises the code, and the IR is left unchanged (Appendix~\ref{app:refine-prompt}).

\section{Experimental Setup}
\label{sec:exp}
\label{sec:exp-setup}

\subsection{Benchmarks and traces}
\label{sec:exp-data}

\paragraph{\tautwobench.} This public customer-support benchmark provides tools that act on a simulated environment \citep{tau2bench}. We use the telecom domain's official split of $74$ training and $40$ test inputs across three task families (MMS, mobile data, and service issues) in \emph{solo mode}, without a simulated user (Appendix~\ref{app:dataset}). We collect eight agent runs per training input. Unless noted, traces come from Sonnet 4.6 with native thinking, which returns its reasoning alongside tool calls. Construction uses only successful traces and Sonnet 4.6 in all three stages. Five-fold cross-validation on the training inputs selects configurations, including $k$; the main comparisons use $k{=}4$. Test inputs are never used for trace generation, configuration selection, or candidate checks.

\paragraph{SOP-Bench.} This public benchmark pairs industrial SOPs with tools and labeled task inputs \citep{sopbench}. We admit $8$ of its $14$ domains (Appendix~\ref{app:dataset} explains the exclusions). The $8$ procedures use $3$--$10$ tools and SOPs of $4$--$12$k characters, and each asks for a decision such as escalating or resolving a case. SOP-Bench has no official split, so we split each procedure in half, stratified by label, into $758$ training and $744$ test inputs. Traces come from eight runs of Sonnet 4.6, as a two-pass ReAct agent, on each training input. We develop with five-fold cross-validation on the training inputs: in each fold, a method that uses traces draws three samples of $k$ successful traces from the other four folds, builds one workflow per sample with Sonnet 4.6, and is scored on the held-out fold, for $k\in\{1,2,4,8,16\}$. Table~\ref{tab:main} uses $k{=}16$, the best $k$ in cross-validation (or within $0.1$ points of it) for every method. For the test, we regenerate the workflows at the $k$ selected in cross-validation, from a new set of agent runs on the training inputs, and run them on the $744$ test inputs; the agent runs three times on each test input. Intervals resample test inputs within each procedure and do not reflect variation across the trace samples used to generate the workflows.

\paragraph{\dset.} This proprietary dataset contains $79$ production procedures, sampled by customer usage from our platform (Section~\ref{sec:intro}). They use $23$ tool types and range from three-step extractions to multi-tool reviews with conditional steps. Each procedure has $20$ construction inputs and up to $20$ disjoint test inputs. We split procedures into Low, Med, and High complexity using only the procedure text and input/output definitions (Appendix~\ref{app:dataset}). Traces come from the platform's two-pass ReAct agent (Sonnet 4.0), which explains each step before calling a tool. Every method uses the same $k{=}3$ traces per procedure and generates one workflow. For \name, extraction and aggregation use Opus 4.8, and the platform generates the final workflow graph with a model of its choice.

\subsection{Compared methods}

All methods receive the procedure and tool definitions. The \emph{original agent}, which produced the traces, is the reference for task outcomes and execution costs. \emph{Zero-shot} gets only these, \emph{few-shot} also gets $k$ full traces, and \emph{execution-only} gets the same traces with the reasoning removed; all three use \name's generation prompt but, unlike \name, keep their first valid workflow. Zero-shot and few-shot run on all three datasets. On \tautwobench and SOP-Bench, we also evaluate AWM \citep{awm}, which keeps a model that chooses steps at run time, and Skill Factory \citep{skillfactory}, whose generated code we run with inputs supplied directly instead of through its model-based skill selection. On \tautwobench only, we evaluate the Bayesian partial-order method BPOP \citep{bpop}, which learns tool order but cannot produce SOP-Bench's decision output (Appendix~\ref{app:baselines}). Finally, we repair workflows for up to three rounds using failures on training inputs: \name on both public benchmarks and zero-shot on \tautwobench, with the same feedback, repair model, and stopping rule within each benchmark (Appendix~\ref{app:guarantees}).

\subsection{Metrics}

\emph{Build rate} is the fraction of requested constructions that yield a usable workflow: the code must load and, for planner-free methods, pass the planning-call checks. On SOP-Bench, code is regenerated until it parses, so every method builds; these retries count as construction calls. \emph{Task success} is the benchmark's score among built workflows, with execution errors counted as failures. On SOP-Bench an answer counts only if it uses the SOP's output format and every scored field matches the label; this is stricter than the benchmark's own substring match, so our numbers are not comparable to published SOP-Bench results. SOP-Bench results are macro averages over its $8$ procedures. Enterprise \emph{completion} is the fraction of executions the platform marks as successfully finished, averaged over procedures. We also report time per task, model calls for construction and execution, and enterprise cost per task. Confidence intervals ($95\%$) are bootstrap estimates, paired where possible; enterprise differences use procedures that both methods ran (Appendix~\ref{app:guarantees}).

\section{Results and Analysis}
\label{sec:results}
\label{sec:exp-main}
\label{sec:analysis}

\subsection{Task outcomes and repair}
\label{sec:results-outcomes}

\begin{table*}[t]
\caption{\textbf{Main results.} \tautwobench: $k{=}4$, chosen by five-fold cross-validation on the $74$ training inputs; results on the $40$ test inputs. SOP-Bench: $k{=}16$, the best $k$ (${\pm}0.1$) in five-fold cross-validation on the training split; results on the $744$ test inputs (agent: three runs per input). \dset: $k{=}3$, not tuned; results on test inputs disjoint from construction, over the procedures each method could run ($70$--$78$ of $79$). Repair starts from $k{=}2$ workflows on \tautwobench and from $k{=}16$ workflows on SOP-Bench. Times: mean per run (\dset: mean\,/\,median). Model calls: construction\,/\,execution (C\,/\,E), with construction as a lower bound. Bold: best success, time, or cost without a runtime planner; ---: not measured or not applicable.}
\label{tab:main}
\centering
\small
\setlength{\tabcolsep}{4pt}
\adjustbox{max width=\textwidth}{%
\begin{tabular}{lrcrrrcrr}
\toprule
& \multicolumn{4}{c}{\tautwobench} & \multicolumn{4}{c}{SOP-Bench} \\
\cmidrule(lr){2-5}\cmidrule(lr){6-9}
Method & Build \% & \shortstack{Task success \%\\{[95\% CI]}} & \shortstack{Calls\\C\,/\,E} & Time (s) & Build \% & \shortstack{Task success \%\\{[95\% CI]}} & \shortstack{Calls\\C\,/\,E} & Time (s) \\
\midrule
Original agent & --- & 80.6 [52.1, 100.0] & ---\,/\,12.1 & 170 & --- & 85.5 [83.0, 88.0] & ---\,/\,11.0 & 46.6 \\
Zero-shot & 100.0 & 38.8 [30.0, 44.4] & 2\,/\,0 & 1.8 & 100.0 & 67.7 [64.5, 70.9] & 2.1\,/\,0 & 0.27 \\
Few-shot & 91.7 & 70.6 [52.1, 88.9] & 2\,/\,0 & 1.6 & 100.0 & 83.6 [81.2, 86.2] & 2.3\,/\,0 & 0.24 \\
\name & 100.0 & 75.8 [66.1, 85.5] & 11\,/\,0 & \textbf{1.3} & 100.0 & 80.6 [78.1, 83.1] & 17.0\,/\,0 & 0.29 \\
\name\ $+$ repair & 94.4 & \textbf{81.2} [74.4, 88.1] & 17.8\,/\,0 & \textbf{1.3} & 100.0 & \textbf{87.8} [86.2, 89.4] & 35.8\,/\,0 & 0.40 \\
\midrule
AWM & 100.0 & 82.3 [68.0, 100.0] & 1\,/\,26.2 & 89.8 & 100.0 & 84.5 [82.0, 86.8] & 1.0\,/\,7.1 & 25.8 \\
Execution-only & 94.4 & 67.9 [53.3, 82.6] & 2\,/\,0 & 1.7 & 100.0 & 80.9 [78.2, 83.6] & 1.9\,/\,0 & 0.24 \\
Skill Factory & 72.2 & 62.3 [44.4, 80.2] & 3\,/\,0 & 2.9 & 100.0 & 60.5 [57.9, 63.2] & 4.4\,/\,0 & \textbf{0.22} \\
BPOP & 100.0 & 30.6 [12.1, 48.4] & 0\,/\,0 & 4.3 & \multicolumn{4}{c}{\emph{not evaluated on SOP-Bench}} \\
\bottomrule
\end{tabular}}
\end{table*}

\begin{table}[t]
\caption{\textbf{\dset results} from Table~\ref{tab:main}, shown separately for readability.}
\label{tab:dset-main}
\centering
\small
\setlength{\tabcolsep}{3pt}
\adjustbox{max width=\linewidth}{%
\begin{tabular}{lccccrr}
\toprule
& \multicolumn{4}{c}{Completion \% [95\% CI]} & \multicolumn{2}{c}{Per task} \\
\cmidrule(lr){2-5}\cmidrule(lr){6-7}
Method & All & Low & Med & High & Time (s) & Cost (\$) \\
\midrule
Original agent & 96.8 [93.5, 99.0] & 95.4 [87.4, 100.0] & 98.5 [97.2, 99.6] & 96.7 [92.4, 99.4] & 153\,/\,68 & 0.127 \\
Zero-shot & 82.7 [73.9, 90.7] & 83.9 [69.6, 95.4] & 80.0 [64.0, 96.0] & 84.0 [68.7, 96.2] & 134\,/\,42 & 0.072 \\
Few-shot & 88.6 [81.1, 95.3] & 91.5 [80.4, 99.3] & 88.6 [75.0, 100.0] & 85.0 [69.3, 98.6] & \textbf{105}\,/\,42 & \textbf{0.033} \\
\name & \textbf{93.8} [88.4, 98.2] & \textbf{96.3} [88.9, 100.0] & \textbf{95.8} [87.6, 100.0] & \textbf{89.2} [78.6, 97.6] & 163\,/\,\textbf{37} & 0.096 \\
\bottomrule
\end{tabular}}
\end{table}

\paragraph{Main comparison.} On \tautwobench, \name reaches $75.8\%$ task success with a $100\%$ build rate (Table~\ref{tab:main}). This is the highest mean among unrepaired methods without a runtime planner. The paired gain over zero-shot is $37.1$ points [$19.8,51.1$]; the margins over few-shot ($5.3$) and execution-only ($7.9$) are not significant (Appendix~\ref{app:guarantees}). The original agent ($80.6\%$) and AWM ($82.3\%$) score higher but call an LLM at every step. On SOP-Bench, \name ($80.6\%$) trails few-shot ($83.6\%$), is level with execution-only ($80.9\%$), and is ahead of zero-shot ($67.7\%$) and Skill Factory ($60.5\%$). AWM ($84.5\%$) and the agent ($85.5\%$) score higher, gaining mostly where a step must read free text at run time. Only \name selects among candidates; with the same selection, few-shot reaches $84.0\%$ in cross-validation (Appendix~\ref{app:baselines}). On some procedures the labels follow a rule the SOP does not state; the IR follows the SOP, so its workflows miss those labels (Appendix~\ref{app:sopbench}). On \dset, \name's mean completion is $93.8\%$, against $96.8\%$ for the agent (paired difference $-3.0$ [$-8.0,1.0$]) and $88.6\%$ for few-shot; the agent's rate overstates its web-procedure success (Section~\ref{sec:exp-crawl}).

\begin{figure}[t]
\centering
\includegraphics[width=0.5\linewidth]{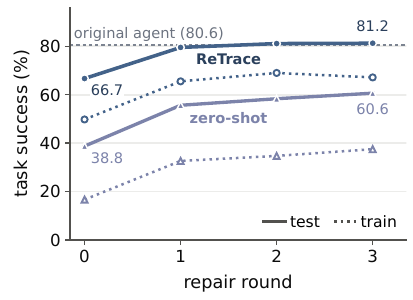}
\caption{\textbf{Repair on \tautwobench.} Test success after each repair round from $k{=}2$ workflows (the unrepaired \name row of Table~\ref{tab:main} uses $k{=}4$). Solid: test; dotted: training inputs, which are harder (Appendix~\ref{app:guarantees}).}
\label{fig:refine}
\end{figure}

\paragraph{Repairing workflows with training feedback.} Starting from $k{=}2$ workflows, three repair rounds raise \name on \tautwobench from $66.7\%$ to $81.2\%$ test success (Figure~\ref{fig:refine}), close to the agent's $80.6\%$, though the interval is too wide to claim parity; zero-shot rises from $38.8\%$ to $60.6\%$. On SOP-Bench, repair raises \name at $k{=}16$ from $80.6\%$ to $87.8\%$, the best result in Table~\ref{tab:main}. There the feedback shows each failed input's label, so the comparison with unrepaired methods is not like-for-like.

\paragraph{Generating workflows from full traces.}
\label{sec:results-full-traces}
On \tautwobench, supplying more full traces to the workflow generator first helps, then lowers build-adjusted cross-validation success, and the build rate falls from $100\%$ at $k{=}16$ to $31\%$ at $k{=}32$. On \dset, nine full traces lower completion to $81.4\%$, below zero-shot's $82.7\%$ (Appendix~\ref{app:raw-traces}). Short SOP-Bench traces ($2$--$8$ steps) show no such decline even at $k{=}16$ (Table~\ref{tab:sopbench-k} in Appendix~\ref{app:sopbench}).

\subsection{What does the agent's reasoning add?}
\label{sec:ana-ablation}

\begin{figure}[t]
\centering
\begin{subfigure}[t]{0.36\linewidth}
  \centering
  \includegraphics[width=\linewidth]{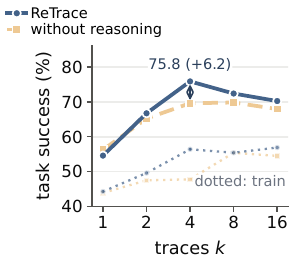}
  \caption{\tautwobench number of traces}
  \label{fig:ablation-k}
\end{subfigure}
\hspace{0.06\linewidth}
\begin{subfigure}[t]{0.36\linewidth}
  \centering
  \includegraphics[width=\linewidth]{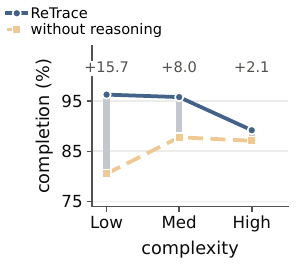}
  \caption{\dset complexity}
  \label{fig:ablation-tier}
\end{subfigure}
\caption{\textbf{Effect of removing the agent's reasoning.} (a) \tautwobench task success by number of traces $k$; dotted: training inputs. (b) \dset completion by complexity group; gray bars: gain from reasoning.}
\label{fig:ablation}
\end{figure}

As an ablation, we keep \name's pipeline and remove the reasoning from every trace step. At $k{=}4$, reasoning adds $6.2$ points of \tautwobench success [$0.5,12.5$]. At $k{=}1$ the version without reasoning is $1.7$ points ahead, but full \name leads at every larger $k$ (Figure~\ref{fig:ablation}a). On \dset, reasoning raises completion from $85.0\%$ to $93.8\%$, a gain of $8.8$ points [$2.5,15.9$] over $77$ procedures. SOP-Bench does not show the effect: without reasoning, \name scores $82.3\%$ against $80.6\%$ on the test inputs at $k{=}16$ (Table~\ref{tab:sopbench-sop} in Appendix~\ref{app:sopbench}). Its short, lookup-like procedures leave the reasoning little to add. Most of the enterprise gain comes from four procedures where the workflow built without reasoning fails on every input, each time at the same code step (Figure~\ref{fig:ablation}b; Appendix~\ref{app:reasoning-details}).

The reasoning also names actions no trace executed: without it, unwitnessed branches citing reasoning fall from $44$ to $3$ at $k{=}4$ (Appendix~\ref{app:unwitnessed}).

\subsection{Execution time and operating cost}
\label{sec:results-cost}

\paragraph{Public benchmarks.} On \tautwobench, \name averages $1.3$\,s per task, $130\times$ faster than the original agent ($170$\,s); AWM takes $90$\,s. In a separate timing run by task depth, \name's workflows stay under $2.5$\,s even on the deepest tasks, where the agent needs $214$\,s (Figure~\ref{fig:efficiency}a in Appendix~\ref{app:results-details}). The workflows call no models, since the tools are simulated. Construction makes at least $11$ model calls, once; the agent's $12.1$ planning calls and AWM's $26.2$ recur on every run. On SOP-Bench, planner-free workflows take $0.22$--$0.40$\,s per task, against $25.8$\,s and $7.1$ model calls for AWM and $46.6$\,s and $11.0$ calls for the agent.

\paragraph{Enterprise costs.} \name lowers mean cost per task by $24\%$, from \$0.127 to \$0.096, counting model calls made inside tools. Few-shot costs less (\$0.033), but its completion is $5.2$ points lower. Median time falls from $68$\,s to $37$\,s ($46\%$), but mean time rises from $153$\,s to $163$\,s (Section~\ref{sec:exp-crawl}).

\subsection{Why enterprise mean time increases}
\label{sec:exp-crawl}

Twenty of the $79$ enterprise procedures use web automation. Without them, \name averages $58$\,s per task against the agent's $92$\,s, at $19\%$ of its cost (Figure~\ref{fig:efficiency}b in Appendix~\ref{app:results-details}). A successful web-tool call takes a median of $270$\,s and a failed one a few seconds, so the agent is sometimes faster only because its calls fail. Web-tool calls succeed $86.6\%$ of the time for \name and $27.3\%$ for the agent. Some agent runs with no successful tool call still count as complete (Appendix~\ref{app:web-failures}).

\subsection{Lessons for practitioners}

First, check tool outcomes, not only completion flags (Section~\ref{sec:exp-crawl}). Second, test the model that writes the workflow: it matters more than the extraction model, and a larger model did not help (Appendix~\ref{app:model-choices}). Third, when traces are long, give the generator few of them: more full traces eventually hurt (Section~\ref{sec:results-full-traces}).

\section{Conclusion}
\label{sec:conclusion}

\name turns an SOP and a few logged agent runs into a workflow in which code, not an LLM, picks each step. The agent's reasoning adds $6.2$ points of task success on \tautwobench and $8.8$ points of enterprise completion, but not on SOP-Bench's short procedures. The workflows cut \tautwobench time per task from $170$\,s to $1.3$\,s and enterprise cost by $24\%$. For high-volume procedures, compiling an agent into a workflow can pay off.

\section{Limitations}
\label{sec:limitations}

\name relies on the agent's reasoning, which may be wrong or may not reflect why the agent acted. Our construction checks confirm that generated code loads and contains no planning calls and, where re-execution is available, count how many logged steps it reproduces; they do not show that the recovered conditions are correct. We evaluate new inputs to the same procedures, not transfer to new procedures, and a change to an SOP or its tools may require rebuilding the workflow. Poor or incomplete traces limit what can be recovered. Enterprise completion is reported by the platform, which can mark failed runs as complete (Section~\ref{sec:exp-crawl}); a human audit of outcomes would give a more reliable measure. The public evaluation covers only the telecom domain of \tautwobench and eight SOP-Bench domains. SOP-Bench results use one workflow generation per trace sample, and its repair feedback includes training labels. Aggregation lets the SOP override traces that contradict it (Section~\ref{sec:results-outcomes}); revising a rule when its cited evidence disagrees is left to future work. We also draw traces at random, without regard to which branches they cover, although the draw matters: results on a single SOP-Bench procedure vary by as much as $50$ points across trace samples. Mixing successful and failed traces tends to help on \tautwobench, though not significantly (Appendix~\ref{app:trace-composition}), and with long traces, full traces hurt beyond a dataset-dependent $k$. We choose $k$ by cross-validation among a few small values ($k\le4$ on \tautwobench), since each trace costs a live agent run. Choosing traces that cover different branches and outcomes, and setting $k$ from trace length, could give a complete IR from fewer agent runs and a workflow that succeeds on its first build, without repair. Future work should also verify that recovered conditions hold before the actions they justify and should cover more public domains.

\section{Ethical Considerations}
\label{sec:ethics}

\paragraph{Public benchmarks.} \tautwobench and SOP-Bench provide publicly available task data and evaluation code \citep{tau2bench,sopbench}. Evaluation on these benchmarks uses existing task data and test environments, without collecting new data from human participants or interacting with real customers.

\paragraph{Enterprise data.} Agent traces can contain sensitive information in task inputs, tool results, and explanations. We report only aggregate enterprise results and do not name the platform.

\paragraph{Deployment.} Reusing a workflow can repeat an incorrect condition or an unfair decision across many cases. The agent's reasoning and written procedures may preserve biases from the original agent or business process; code-based step selection does not remove those biases. Workflow owners should review recovered conditions, restrict tool access, and require human review for actions with significant consequences for users. Access restrictions also matter because the same automation could scale unauthorized collection or use of sensitive data. Our evaluation does not assess differences in error rates across user groups.

{\small
\bibliographystyle{plainnat}
\bibliography{references}
}

\clearpage
\newpage

\appendix
\begin{center}
  {\Large\bfseries Appendix}
\end{center}
\section{Dataset details}
\label{app:dataset}

\paragraph{\tautwobench.} We use only the telecom domain. The airline and retail environments do not support the solo mode that a planner-free workflow needs, and only telecom provides a written workflow document for each task family. A benchmark is useful here only if the original agent solves its tasks often enough to leave a procedure worth compiling. We pin the benchmark code to a fixed revision and keep its official split of $74$ training and $40$ test inputs across three task families. Traces come from running the original agent eight times on each training input; no trace comes from a test input. We vary the number of traces over $k\in\{1,2,4,8,16\}$, limited to the values each family supports, and add $k{=}32$ for the full-trace sweep (Appendix~\ref{app:raw-traces}). Because solo mode removes the simulated user, our \tautwobench scores are not comparable to published results.

\paragraph{SOP-Bench.} We pin the public SOP-Bench code to a fixed revision \citep{sopbench} and admit $8$ of its $14$ domains (Table~\ref{tab:sopbench-data}). Two excluded domains (aircraft inspection, patient intake) have tools that return the answer columns directly, so there is nothing to induce. Three (content flagging, order fulfillment, warehouse package inspection) declare no input columns, so the benchmark's loader passes the whole row, label included, as the task. In the last (video annotation), $20$ of $26$ tools are empty stubs. We drop one \texttt{dangerous\_goods} row whose required input is missing. SOP-Bench has no official split, so we split each domain $50/50$, stratified by label, and freeze the split before collecting traces. We called every tool on every row and confirmed that no tool returns a row's label. Two adaptations apply to every method: tool results are normalized (whitespace in keys becomes an underscore, and a single wrapping object is removed), and an answer given only as a structured payload is rendered into the SOP's output tag before scoring. For \name only, the workflow generator is told which task-input field each copied tool argument comes from, because its argument tracing otherwise sees only earlier tool results. The strict score counts an answer only if it appears in the SOP's output format and every scored field matches; SOP-Bench's own scorer accepts substring matches, such as ``do not ESCALATED'' for ``ESCALATED''. Traces come from Sonnet 4.6 as a two-pass ReAct agent, eight runs per training input ($6{,}064$ traces).

\begin{table}[!tbhp]
\caption{\textbf{Admitted SOP-Bench domains.} Majority: share of the most frequent training label. Agent: the original agent's success over its training runs.}
\label{tab:sopbench-data}
\centering
\small \setlength{\tabcolsep}{3pt}
\adjustbox{max width=\linewidth}{%
\begin{tabular}{lrrrrrr}
\toprule
Domain & Train & Test & Tools & SOP (k chars) & Majority \% & Agent \% \\
\midrule
customer\_service & 79 & 77 & 10 & 7.6 & 35.4 & 95.7 \\
dangerous\_goods & 137 & 136 & 4 & 4.1 & 32.8 & 69.8 \\
email\_intent & 93 & 93 & 5 & 4.0 & 25.8 & 96.8 \\
know\_your\_business & 45 & 45 & 8 & 7.3 & 37.8 & 53.6 \\
referral\_abuse\_v1 & 101 & 99 & 3 & 7.3 & 52.5 & 100.0 \\
referral\_abuse\_v2 & 102 & 98 & 6 & 11.9 & 65.7 & 98.2 \\
traffic\_spoofing & 101 & 99 & 6 & 4.2 & 45.5 & 78.2 \\
video\_classification & 100 & 97 & 10 & 5.9 & 51.0 & 84.4 \\
\midrule
Total / mean & 758 & 744 & & & 43.3 & 84.6 \\
\bottomrule
\end{tabular}}
\end{table}

\paragraph{Enterprise dataset (\dset).} \dset is a private collection of production procedures from the platform described in Section~\ref{sec:intro} (Table~\ref{tab:dset}). Each procedure is written as a structured Markdown document with explicit branches and typed JSON input and output schemas. We sampled the procedures by customer usage from a much larger set, and they cover several business areas. Each procedure has two disjoint input pools, one for traces and one for testing, so no method sees a test input during construction. Because \dset is private, we report only aggregate results that do not identify its source.

For the web-tool results in Section~\ref{sec:exp-crawl}, a tool call counts as successful only when its own output reports success, not when the run ends. The platform statistics in Section~\ref{sec:intro} come from the platform's automation registry and execution log from April 1 to July 1, 2026. They cover $5{,}115$ new automations ($4{,}694$ agents that plan at run time and $421$ fixed workflows that never ask a model for the next step), together with the execution counts from which we take the ten most frequently executed automations.

\begin{table}[!htbp]
\caption{\dset at a glance. \emph{Top}: scale, procedure size ($p_{50}$\,/\,$p_{95}$), and the original agent's time per task, which is long-tailed because of the web-automation procedures. \emph{Bottom}: procedures in which the original agent called each tool type on the test inputs; a procedure can use several types, so the column sums to more than $100\%$.}
\label{tab:dset}
\centering
\small
\begin{tabular}{lr}
\toprule
Property & Value \\
\midrule
Procedures                             & $79$ \\
Test inputs per procedure              & ${\le}20$ \\
Steps per procedure                    & $5$\,/\,$14$ \\
Tool types per procedure               & $4$\,/\,$8$ \\
Original-agent time (s/task)           & $68$\,/\,$729$ \\
\bottomrule
\end{tabular}

\vspace{6pt}
\begin{tabular}{lrr}
\toprule
Tool type & Procedures & \% \\
\midrule
LLM invocation                & $45$ & $57\%$ \\
Web browsing                  & $20$ & $25\%$ \\
Code interpreter              & $14$ & $18\%$ \\
Image and table QA            & $6$ & $8\%$ \\
Ticket and case retrieval     & $6$ & $8\%$ \\
Optical character recognition & $5$ & $6\%$ \\
Text classification           & $3$ & $4\%$ \\
Document conversion           & $2$ & $3\%$ \\
Other                         & $7$ & $9\%$ \\
\bottomrule
\end{tabular}
\end{table}

\paragraph{Complexity groups.} We group procedures by a difficulty score computed \emph{only} from the procedure text and its input and output schemas, never from execution results, so the grouping does not favor any method. The score is the mean of two standardized components. The \emph{structural score} averages the standardized procedure length, step count, branch count, number of tool types, and number of schema fields. The \emph{rubric score} is a model's $0$--$5$ rating of ambiguity, branching, data handling, error handling, and step count. The two components are related but not redundant (Spearman $0.62$ over the $79$ procedures). We split the combined score into thirds: Low ($27$ procedures, score $[-2.03,-0.39]$), Med ($26$, $[-0.37,+0.32]$), and High ($26$, $[+0.33,+1.46]$). Every structural measure increases from Low to High, though steps per tool type barely changes from Low to Med (Table~\ref{tab:tiers}).

\begin{table}[!htbp]
\caption{\textbf{Structure of the \dset complexity groups.} Means per procedure.}
\label{tab:tiers}
\centering
\small
\begin{tabular}{lrrr}
\toprule
Measure & Low & Med & High \\
\midrule
Tool types                  & $2.8$ & $4.3$ & $5.8$ \\
Steps                       & $3.3$ & $5.2$ & $8.5$ \\
Branches                    & $5.5$ & $7.8$ & $12.2$ \\
Procedure length (words)    & $730$ & $873$ & $1{,}134$ \\
Steps per tool type         & $1.20$ & $1.22$ & $1.47$ \\
\bottomrule
\end{tabular}
\end{table}

\paragraph{Example procedures.} Because \dset is private, we cannot show its procedures. The two below are \emph{synthetic}: we wrote them in \dset's Markdown format and tool vocabulary, with neutral tasks in place of business content. They show the kind of steps and branches a method must recover.

\begin{artifact}{Synthetic procedure 1: message triage}
# SOP: Triage an incoming support message
## Goal
Classify the message, summarize it, and decide whether a human is needed.
## Inputs
- messageText (string): the raw incoming message.
## Steps
### Step 1: Classify
- Call the text model to label the message as one of
  {billing, technical, account, other}.
### Step 2: Summarize
- Produce a one-sentence summary of the request.
### Step 3: Escalation check
- If the label is "technical" and the message reports data loss,
  set needs_human = true; otherwise needs_human = false.
## Output
- category (string), summary (string), needs_human (bool).

input schema : {"messageText": "string"}
output schema: {"category": "string", "summary": "string",
                "needs_human": "boolean"}
\end{artifact}

\begin{artifact}{Synthetic procedure 2: form extraction}
# SOP: Extract fields from a scanned form
## Goal
Read a scanned form and return its key fields, flagging low-confidence reads.
## Inputs
- fileReference (string): handle to the uploaded file.
## Steps
### Step 1: Convert
- Convert the file to a text-extractable format.
### Step 2: Read
- Run optical character recognition to obtain the page text.
### Step 3: Extract
- Call the text model to pull {name, date, total} from the page text.
### Step 4: Confidence gate
- If any field is empty or OCR confidence < 0.8,
  set review_required = true; otherwise false.
## Output
- fields (object), review_required (bool).

input schema : {"fileReference": "string"}
output schema: {"fields": "object", "review_required": "boolean"}
\end{artifact}

\FloatBarrier
\section{Detailed workflow construction}
\label{app:method-construction}

\begin{algorithm}[!htbp]
\small
\caption{Detailed construction procedure for Python workflows.}
\label{alg:retrace-detail}
\KwIn{the SOP $P$ and traces $\mathcal{T}_k$, each $\tau=(s_1,\dots,s_L)$ with $s_t=(r_t,a_t,v_t,o_t)$}
\KwOut{a workflow $\pi$ that chooses steps without calling an LLM}
\SetKwProg{Stage}{Stage}{:}{end}
\Stage{\textnormal{1. \textsc{Extract}} $s_t \mapsto z_t$, one step at a time}{
  \ForEach{$\tau^{(i)}\in\mathcal{T}_k$ and each $s_t\in\tau^{(i)}$, in parallel}{
    $z^{(i)}_t \gets (g_t,\alpha_t,\;a_t,v_t,\omega_t)$ read off $s_t$\;
  }
  \Return{one record per step; nothing is merged or compared yet}
}
\BlankLine
\Stage{\textnormal{2. \textsc{Aggregate}} $\{z_t\}\times P \mapsto \mathcal{I}$}{
  \ForEach{$i\le k$, in parallel}{
    summarize trace $i$'s records against $P$\;
  }
  $\mathcal{I} \gets$ joint merge of the first $\min(k,8)$ summaries with $P$\;
  \For{$i=9$ \KwTo $k$}{
    $\mathcal{I} \gets$ add trace $i$'s summary to $\mathcal{I}$, keeping all existing elements\;
  }
  \Return{$\mathcal{I}=\big(\chi,\langle(g_1,\alpha_1),\dots,(g_m,\alpha_m)\rangle\big)$ once it passes the consistency check}
}
\BlankLine
\Stage{\textnormal{3. \textsc{Synthesize}} $\mathcal{I}\mapsto\pi$}{
  \For{$c=1$ \KwTo $3$}{
    generate a candidate from $\mathcal{I}$, with at most one fix call on its own errors\;
    discard it if it fails the static check or makes a planning call when re-executed\;
  }
  \If{no candidate remains}{\Return{construction failure}}
  \Return{the remaining candidate that reproduces the most recorded steps of $\mathcal{T}_k$}
}
\end{algorithm}
Algorithm~\ref{alg:retrace-detail} gives the full construction procedure summarized in Section~\ref{sec:method-retrace}. Generating several candidates and screening the code apply only to the public benchmarks; the enterprise platform builds one workflow graph without re-executing candidates (Appendix~\ref{app:guarantees}). Extraction reads each step independently. Aggregation first summarizes each trace against the SOP and then merges the summaries; Appendix~\ref{app:aggregation-details} describes how more than eight traces are handled. The merge checks that earlier rules are kept or removed with a reason and that every alternative named in the reasoning is accounted for. These checks concern the recorded evidence, not whether the rules are correct.

\section{Prompts}
\label{app:prompts}

Traces can come from agents with native thinking, which return their reasoning alongside each tool call, or from two-pass ReAct agents \citep{react}, which write an explanation before each call. Both provide the step fields $s_t=(r_t,a_t,v_t,o_t)$ of Section~\ref{sec:method-setup}. The three stages of Section~\ref{sec:method-retrace} use five kinds of model calls: extraction once per trace step; a per-trace summary and a cross-trace merge, which together make up aggregation; and workflow generation with at most one fix call. We wrote the prompts for \tautwobench and used them on \dset and SOP-Bench without changes (apart from the SOP-Bench input-field note in Appendix~\ref{app:dataset}), including their example values, so a few \tautwobench tool and argument names appear in them as format examples. Because nothing was re-tuned, \dset also tests whether the extraction and aggregation prompts transfer; on \dset every method uses the platform's own generation prompt. Apart from the trace-loading code, our implementation contains no tool names from \dset or SOP-Bench. We summarize each prompt below.

\paragraph{Models for each stage.} On \tautwobench all three stages use Sonnet 4.6. Appendix~\ref{app:model-choices} changes the extraction or synthesis model while keeping the others fixed; we never change the aggregation model. On \dset, the traces come from a Sonnet 4.0 agent, extraction and aggregation use Opus 4.8, and the platform generates the workflow with its own prompt and model, which we do not control.

\paragraph{Sampling the merge.} We run the merge several times rather than once: five independent runs on the same evidence for $k\le2$ and three for larger $k$. We measure agreement between runs by the tool calls each rule contains rather than by its wording, and take the most common structure as the base. A rule that only some runs contain is kept as an open item; regenerating the whole document instead worked worse in our tests. When a single new trace is added to an existing IR, we run the merge once, since the existing IR already serves as the base.

\paragraph{Extraction (Stage 1).} This prompt turns one trace step into a record. The model replaces case-specific details (names, IDs, quoted values) with typed variables and reports only decision logic that carries over to new inputs: the situation the step responds to, the action it takes, the observable condition that justifies the action, and whether the step ends the task. Each field must be grounded in the tool called and the result just received, and a condition may use only information available before the action. The reasoning-side fields are \emph{precondition} (the condition $g_t$), \emph{action} (the intended action $\alpha_t$), \emph{state}, and \emph{is\_terminal}; the tool-side fields are the tool, its arguments with their JSON types, and the source of each argument.

\paragraph{Per-trace summary (Stage 2).} For a single trace, this prompt records which procedure steps the trace visited, under what conditions, with what arguments and results, which steps it skipped, and where the run went wrong and how it ended.

\paragraph{Merge and IR format (Stage 2).} This prompt combines all per-trace summaries and the recorded tool results into the IR, a document with a fixed format. It lists the opening steps, the completion check, and an ordered list of \emph{decision rules}, each pairing a condition with the action it selects and the evidence for it. While the completion check fails, the first rule whose guard holds runs. The opening steps are a printing convention, not a separate component: they are the rules whose conditions hold before any tool has been called. The document also marks \emph{unwitnessed} branches that no trace reached, named either by the SOP or by a step's reasoning, and lists material that was explicitly excluded. Every element cites the trace steps or SOP statements that support it. When a new trace is added to an existing IR, the prompt must keep every existing rule, or remove it with a stated reason, and may confirm, narrow, extend, or contradict it.

\paragraph{Workflow generation and consistency check.} The generation prompt turns the IR into a workflow with a single decision function and forbids model calls that choose steps during execution. Before generation, a check uses rule names and citations to confirm three things, without judging what the rules mean: every rule from the summaries is kept or explicitly removed, every merged step has a citation, and every alternative named in the reasoning appears either as a branch or as cited excluded material.

\paragraph{Repair from training failures.}
\label{app:refine-prompt}
Repair starts from a built workflow (round $0$) and runs it on a fixed set of training inputs of the same task family ($4$ on \tautwobench, $16$ on SOP-Bench). If any input fails, the repair model receives one prompt and returns a complete replacement workflow. The new workflow passes the same static checks as the first; when it fails them, the model gets up to two more attempts with the static-check error, separate from the repair rounds. The loop stops when every repair input succeeds, after three revisions, or when a revision cannot be built. It keeps the round with the highest score on the repair inputs, preferring the earlier round on ties. Validation and test inputs never enter the loop. The repair prompt receives neither the IR nor the traces, so repair changes only the code, and the IR remains the IR used for round $0$. Infrastructure faults stop construction instead of being shown as failures. The system prompt is the workflow-generation system prompt; the user prompt is a single JSON object, in which ``refinement'' and ``program'' mean repair and workflow:

\begin{artifact}{Repair prompt (structure; fields abbreviated)}
Repair the previous complete Python policy using only the visible SOP,
declared runtime/tool surface, and structured failures from the designated
train refinement tasks. Preserve correct behavior, address the listed
failures, and return one complete replacement program.

{ "sop": <procedure text>,
  "runtime": <decide(obs) interface, budget>,  "tools": <tool schemas>,
  "refinement": {
    "round_index": r, "max_rounds": 3, "n_refine": n,
    "previous_program": <full source of round r-1>,
    "requirements": { "policy_model_calls": 0,
                      "return_complete_program": true, ... },
    "feedback": { "completed_round": r-1, "primary_score": <share solved>,
      "failures": [ { "task": <inputs>, "status": <completed | defect>,
                      "score": 0, "failure_detail": "<Type>: <message>",
                      "steps": [ <observation, action, tool results> ],
                      "answer_check": { "answer": ..., "label": ...,
                                        "field_matches": ... } } ] } } }
\end{artifact}

Only failed inputs appear in the feedback. The failure detail gives an exception's type and message but no traceback; a workflow that exceeds its step budget receives a fixed message instead. On \tautwobench each step records the observation, the action, its results, and whether the run ended, and there is no \texttt{answer\_check}: the model sees only the score and the execution record. On SOP-Bench the steps record actions and results, with long text cut at $4{,}000$ characters, and \texttt{answer\_check} shows the workflow's answer, the training label, and which fields match. A previous workflow longer than $64$k characters is shortened to its first and last $32$k.

\FloatBarrier
\section{Construction guarantees and approximations}
\label{app:guarantees}

\paragraph{Repeated construction and build rates.} For each configuration that uses traces on \tautwobench, we sample $k$ traces three times and generate four workflows independently from each sample. Over three task families this gives $36$ requested constructions ($3$ samples $\times$ $4$ generations $\times$ $3$ families). Repeated generation captures the variation between model-written workflows. Zero-shot uses no traces and so requests only $12$ constructions. Build rates in Table~\ref{tab:main} therefore move in steps of $2.8$ points for methods that use traces; the training folds contain more constructions and give finer steps. We track requested, built, executed, and scored workflows separately and report exclusions at each stage.

\paragraph{Confidence intervals.} Confidence intervals come from a deterministic paired bootstrap ($2000$ resamples, seed $0$), blocked by family, fold, and trace sample. Except as noted next, every \tautwobench difference we report with an interval is this paired estimate over the $120$ matched cells, not the difference between the two values in the table. Zero-shot uses no traces, and AWM, the original agent, and the repair runs were evaluated separately; none of them shares trace samples with the other methods, so their comparisons are paired by family and test input over the $40$ test inputs instead. All \tautwobench results pool the three trace samples over the same $40$ test inputs, so they are on one scale. Because the paired estimator weights test inputs equally, a paired difference need not equal the difference of the table means. We do not correct for multiple comparisons. With $40$ test inputs, \name's margins over few-shot, $+5.3$ [$-10.2,+21.7$], and over execution-only, $+7.9$ [$-3.6,+20.5$], are not significant, nor is repair's $+0.6$ [$-19.8,+20.7$] over the original agent. The margin over zero-shot, $+37.1$ [$+19.8,+51.1$], is significant, as are both reasoning-removal effects in Section~\ref{sec:ana-ablation}.

\paragraph{Call counts are lower bounds.} The construction call counts in Table~\ref{tab:main} understate the \name rows (with and without repair), because extraction results are cached and reused, and AWM's, because its construction calls were not fully logged. Three construction stages also do not mean three calls. At $k{=}4$, one \name construction makes an extraction call per trace step, a summary call per trace, several merge calls, a generation call, and at most one fix call. The table counts only the calls that the extraction cache did not serve.

\paragraph{Procedure counts on \dset.} A method's \dset completion is the mean over the procedures for which it produced a runnable workflow, with each procedure weighted equally and scored on its own test inputs. The overall value is therefore not the mean of the three group values, because each method has its own group sizes. Out of $79$ procedures ($27$ Low, $26$ Med, $26$ High), the original agent ran $76$ ($27$, $23$, $26$), both versions of \name $77$ ($27$, $25$, $25$), zero-shot $78$ ($27$, $25$, $26$), and few-shot $70$ ($27$, $22$, $21$). \dset intervals come from a bootstrap over procedures ($2000$ resamples, seed $0$); a paired \dset difference resamples only the procedures that both methods ran, and every \dset difference reported with an interval is this paired estimate.

\paragraph{Zero-shot and few-shot rows.} The zero-shot and few-shot rows of Table~\ref{tab:main} come from a separate evaluation round on the same $40$ test inputs with the same scorer; the original agent's score in that round matches the one reported here exactly. That round kept every scored attempt ($160$ for zero-shot; $432$ for few-shot at $k{=}4$), so their intervals use the same bootstrap as the other \tautwobench rows.

\paragraph{Repair procedure.} Each repair round runs the workflow on the training inputs and uses its failures to revise it. The traces, tasks, feedback, workflow-generation model, and stopping rule are fixed, and test inputs are evaluated only after the last round. Each version depends on the previous repair, so these results do not average independent generations as the main comparison does.

Three rounds raise test success from $66.7\%$ to $81.2\%$ [$74.4,88.1$] for \name and from $38.8\%$ to $60.6\%$ for zero-shot. The estimated test gains are $14.6$ and $21.9$ points, and the gains on the training inputs are $17.4$ and $20.8$ points. Training success stays below test success, likely because the training inputs are harder: the original agent succeeds on $65.0\%$ of its training runs against $80.6\%$ on test (Appendix~\ref{app:source}), and the hardest family, \texttt{mms\_issue}, makes up $33$ of the $74$ training inputs but $16$ of the $40$ test inputs. The repaired workflow's difference from the original agent ($80.6\%$) is $0.6$ points [$-19.8,20.7$], which does not establish equal performance. Both the starting and the repaired \name workflows run without planning calls. We do not run repair on \dset: the platform provisions each workflow, and we currently cannot re-execute candidates offline during construction.

\paragraph{Enforcing no planning calls (public benchmarks).} On \tautwobench and SOP-Bench we enforce the no-planning-call rule in three layers. First, a static check requires that the generated code parses and loads, exposes a single synchronous decision function, imports only the tool actions and a restricted part of the standard library, and contains no reflection, model-provider code, or dynamic evaluation. Second, the code runs in a separate sandbox process whose import hook blocks every model provider and which limits memory and time. Third, each candidate is re-executed on the training inputs, and we reject any candidate that makes a planning call. Code that passes all three layers made no planning call in any execution we observed. On \dset the platform builds a typed workflow graph that we cannot run in our sandbox, so none of the three layers applies; there, the guarantee is that the graph contains no planner node.

\paragraph{Choosing among candidates (public benchmarks).} When candidates can be re-executed, generation produces up to three of them and keeps the one that reproduces the most recorded steps, breaking ties in favor of fewer decision steps. Each later candidate sees the earlier ones and their results on each example, so the choice is a small guided search rather than a single filter. This selection is not available on \dset, where the platform builds the workflow and we cannot re-execute candidates, so we generate one workflow per procedure there. The merge's consistency check needs no execution environment and is used on all three datasets.

\paragraph{Rules for branches absent from the traces.}
\label{app:unwitnessed}
The IR marks a branch as \emph{unwitnessed} when its action never appears in the supplied traces. Its support comes from the SOP or from reasoning that names an alternative action. We count each branch once, at its decision rule; counting every line that carries the mark would inflate the total by about $1.7\times$. A branch is supported by reasoning when it cites a trace step's reasoning rather than only the SOP.

\begin{figure}[!htbp]
\centering
\includegraphics[width=0.6\linewidth]{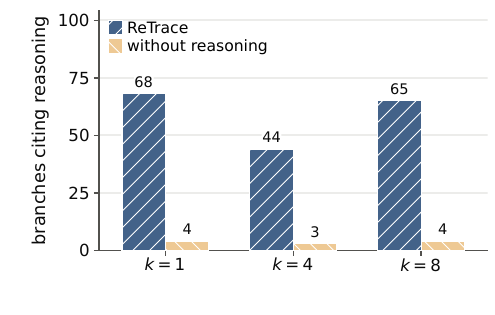}
\caption{\textbf{Unwitnessed branches citing the agent's reasoning} in \tautwobench IRs built with and without reasoning. Counts describe IRs before candidate selection, not branches reached at test time.}
\label{fig:ablation-unwit}
\end{figure}

Across fifteen saved IRs built from $k\in\{1,\dots,16\}$ traces, $23$ of $115$ unwitnessed branches cite trace reasoning; the rest cite only the SOP. In a larger matched set built from the same trace samples, removing reasoning reduces the branches supported by reasoning from $68$ to $4$ at $k{=}1$, from $44$ to $3$ at $k{=}4$, and from $65$ to $4$ at $k{=}8$ (Figure~\ref{fig:ablation-unwit}).

The fifteen IRs come from an earlier construction round with a different code version; the main round saved workflows but not their IRs. The larger matched counts come from saved merge outputs before candidate selection. Neither set records whether test executions reach these branches, so the counts describe construction evidence and do not by themselves explain the task-success gains in Section~\ref{sec:ana-ablation}.

\paragraph{Building the IR without the SOP.} Aggregation takes the SOP $P$ as an input (Section~\ref{sec:method-aggregate}). Without it, the IR fails the same way each time: with no structure to attach steps to, it follows the order of the trace, indexes steps by their position in that run, and copies the sampled case's values into the tool calls. The result is longer than its evidence; on \tautwobench, $24{,}878$ characters from an $8{,}344$-character trace.

\paragraph{Aggregation beyond eight traces.}
\label{app:aggregation-details} Merging all $k$ per-trace summaries at once overflows the context as $k$ grows, and adding them one at a time from the start takes $k$ sequential merge calls. We therefore merge the first eight summaries together and then add each further trace to that base, keeping every existing element and citation. The merge input stays bounded, and the IR grows slowly, to about $10$k tokens at $k{=}16$. Each later trace is compared with the merged IR rather than with every other later trace. The \tautwobench and \dset main results ($k{=}4$ and $k{=}3$) never reach the incremental step; the SOP-Bench results at $k{=}16$ do.

\paragraph{Batch merge versus incremental merge.} Aggregation can merge a fixed set of traces at once (the default) or add each new trace to the IR as it arrives. The two give similar training results: $49.6\%$ [$37.7,62.9$] for the batch merge and $49.8\%$ [$37.6,62.5$] for the incremental merge at $k{=}2$, and $56.4\%$ [$44.7,69.2$] and $54.9\%$ [$41.3,69.4$] at $k{=}4$. The incremental IR also stays compact, at $4{,}601$ tokens and $25$ decision rules at $k{=}2$ and $5{,}921$ tokens and $27$ rules at $k{=}4$. At least up to $k{=}4$, \name can therefore add new runs as a procedure keeps executing, without waiting for a batch.

\paragraph{Argument types.} Extraction reads the tool records together with the reasoning (Section~\ref{sec:method-extract}), so every argument reaches the IR as a typed variable with a recorded source, or marked unknown. The order of actions alone provides neither: it shows which call followed which but leaves each argument's value and type to the workflow generator, and a numeric field filled with a string passes construction and fails during execution. The guarantee ends at the IR: generation is a model call, so whether the workflow respects a type depends on the generation model; \name controls only the information that model receives.

\FloatBarrier
\section{Additional experimental results}
\label{app:results-details}

\begin{figure}[!htbp]
\centering
\begin{subfigure}[t]{0.49\linewidth}
  \centering
  \includegraphics[width=\linewidth]{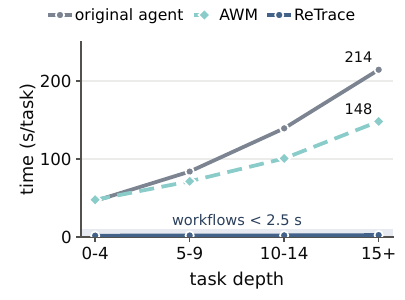}
  \caption{\tautwobench time}
  \label{fig:eff-tau2}
\end{subfigure}
\hfill
\begin{subfigure}[t]{0.49\linewidth}
  \centering
  \includegraphics[width=\linewidth]{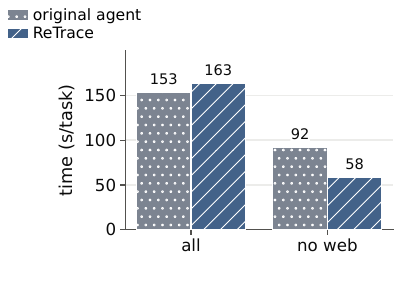}
  \caption{\dset mean time}
  \label{fig:eff-dset}
\end{subfigure}
\caption{\textbf{Execution time per task.} (a) \tautwobench, by task depth (tool calls per task), from a separate timing run that uses \name's $k{=}2$ workflows; the band spans all workflows without a runtime planner. (b) \dset mean time over the procedures each method ran, and over those without web automation.}
\label{fig:efficiency}
\end{figure}

\subsection{Reasoning removal: trace counts, complexity, and cost}
\label{app:reasoning-details}

Removing the reasoning keeps each tool call, its inputs, and its result; all construction stages and models stay the same. On \tautwobench, the version without reasoning is $1.7$ points ahead at $k{=}1$. For every larger number of traces we tested, full \name leads by $1.5$--$6.2$ points, and at $k{=}4$ the gain is $6.2$ points [$0.5,12.5$]. The best version without reasoning reaches $69.8\%$ at $k{=}8$, still below full \name's $75.8\%$ at $k{=}4$ (Figure~\ref{fig:ablation}a).

On \dset, completion rises from $85.0\%$ without reasoning to $93.8\%$ with it, a paired gain of $8.8$ points [$2.5,15.9$] over $77$ procedures. Mean time and cost per task are $178$\,s and \$0.101 without reasoning and $163$\,s and \$0.096 with it, so removing reasoning brings no efficiency benefit. The completion gains are $15.7$ points [$3.7,29.6$] for Low, $8.0$ for Med, and $2.1$ for High complexity procedures; only the Low gain is significant, since only its $95\%$ interval excludes zero (Figure~\ref{fig:ablation}b). The trend comes from four all-or-nothing failures, not from a steady effect of complexity: on four Low procedures, the workflow built without reasoning fails on all $20$ test inputs, each time at the same code step (in one, the code reads a field that is missing), while full \name completes all $20$. Without these four, the Low gain falls to about $1$ point, and $22$ of the $27$ Low procedures show no difference.

\subsection{Enterprise web-tool failures}
\label{app:web-failures}

This section details the web-automation results summarized in Section~\ref{sec:exp-crawl}. On eight of the twenty procedures, the original agent makes no successful web call. It passes the tool an unresolved instruction, receives a rejection, retries two to five times, and then writes an answer without any successful tool result. On these procedures it takes $54$--$85$\,s per execution and records $100\%$ completion with $0\%$ tool-call success. One answer reads \texttt{"Unable to extract title - web extraction failed"}, yet the platform marks the run complete. Across the test set, $184$ agent executions are marked successful although their outcome field says the goal was not achieved, and another $454$ have no outcome field. Neither problem occurs in \name's outputs.

\name's completion falls from $96.3\%$ on Low to $89.2\%$ on High complexity procedures, while the agent's ranges from $95.4\%$ to $98.5\%$ with no clear trend. Platform completion does not directly measure whether the user's goal was met. Tool outcomes add evidence about execution quality, but they do not establish that every workflow is correct.

\subsection{Successful and failed traces}
\label{sec:ana-trace}
\label{app:trace-composition}

We compare workflows built from successful traces, from failed traces, and from a balanced mix of both. The main Sonnet trace pool is $65\%$ successful and contains no failures for one family, so this comparison uses Haiku 4.5 traces, whose $29\%$ success rate gives both outcomes in every family, with $k{=}4$.

\begin{figure}[!htbp]
\centering
\begin{subfigure}[t]{0.48\linewidth}
  \centering
  \includegraphics[width=\linewidth]{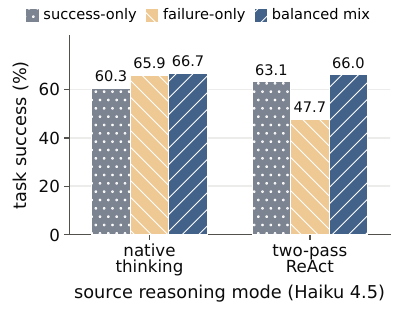}
  \caption{\tautwobench}
  \label{fig:trace-sel-tau2}
\end{subfigure}
\hfill
\begin{subfigure}[t]{0.48\linewidth}
  \centering
  \includegraphics[width=\linewidth]{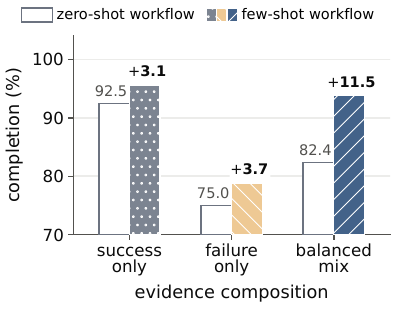}
  \caption{\dset}
  \label{fig:trace-sel-dset}
\end{subfigure}
\caption{\textbf{Which traces should construction use?} (a) Workflows built from Haiku 4.5 traces, $k{=}4$. (b) Enterprise few-shot workflows; each trace mix is compared with zero-shot on its own procedures, so the groups are not controlled comparisons. The y-axis starts at 70.}
\label{fig:trace-sel}
\end{figure}

\paragraph{\tautwobench.} With native thinking, a mix of outcomes gives $66.7\%$ task success against $60.3\%$ for successful traces only; with two-pass ReAct, it gives $66.0\%$ against $63.1\%$ (Figure~\ref{fig:trace-sel}a). Neither paired gain ($6.3$ and $2.9$ points; Appendix~\ref{app:guarantees}) is significant: both 95\% intervals include zero. Relative to successful traces only, failed traces alone raise success by $5.6$ points with native thinking but lower it by $15.4$ points [$-30.3,-0.4$] with two-pass ReAct, so failures alone are not reliably useful.

\paragraph{\dset.} This experiment generates few-shot workflows directly from the platform agent's traces, without \name's extraction and aggregation. Each trace mix (successful, failed, or mixed) is compared with zero-shot on the procedures where that mix is available. The groups contain $8$--$34$ procedures and differ in complexity, because failed traces exist mostly for harder procedures. Completion gains over zero-shot are $11.5$ points for mixed traces, $3.1$ for successful traces, and $3.7$ for failed traces (Figure~\ref{fig:trace-sel}b). Because the groups contain different procedures, these gains are not a controlled ranking. Together, the two experiments suggest trying mixed traces, but they do not show that failed traces are always safe to include alongside successes.

\subsection{Model choices during construction}
\label{sec:ana-model}
\label{app:model-choices}

On \tautwobench with $k{=}4$, we change one of three models at a time: the agent that produces the traces, the extraction model, or the synthesis model, which writes the workflow. All other models stay at Sonnet 4.6.

\begin{figure}[!htbp]
\centering
\begin{subfigure}[t]{0.48\linewidth}
  \centering
  \includegraphics[width=\linewidth]{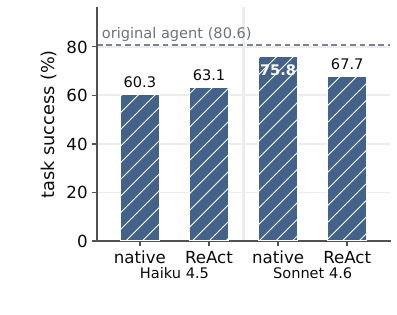}
  \caption{trace-producing agent}
  \label{fig:model-source}
\end{subfigure}
\hfill
\begin{subfigure}[t]{0.48\linewidth}
  \centering
  \includegraphics[width=\linewidth]{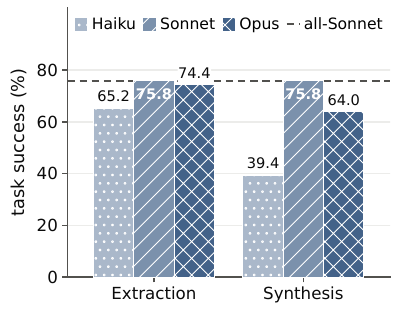}
  \caption{construction stage}
  \label{fig:model-stage}
\end{subfigure}
\caption{\textbf{Model changes on \tautwobench, $k{=}4$.} (a) Changing the agent that produces the traces. (b) Changing the model of one construction stage while the other stages stay at Sonnet 4.6.}
\label{fig:model-iter}
\end{figure}

\paragraph{Agent producing the traces.} Sonnet traces give workflows with $75.8\%$ and $67.7\%$ task success under native thinking and two-pass ReAct, against $60.3\%$ and $63.1\%$ for Haiku traces (Figure~\ref{fig:model-iter}a). Native thinking leads ReAct by $8.1$ points for Sonnet but trails it by $2.8$ for Haiku, so the better reasoning format may depend on the model.

\paragraph{Extraction and synthesis models.} Replacing Sonnet with Haiku lowers success by $36.5$ points in synthesis but by only $10.6$ in extraction (Figure~\ref{fig:model-iter}b; these are paired estimates, which need not equal the differences between bars, Appendix~\ref{app:guarantees}). Replacing Sonnet with Opus 4.8 also lowers the averages, by $11.8$ points in synthesis and $1.5$ in extraction. Synthesis is the more sensitive stage in both cases. The synthesis model should therefore be tested before use, since a larger model is not automatically better. These results do not explain why Opus scores below Sonnet.

\section{SOP-Bench results in detail}
\label{app:sopbench}

All values are strict task success (\%). Table~\ref{tab:sopbench-sop} gives test results at $k{=}16$: each method builds three workflows, one per trace sample from the full training split, and runs each once on every test input; the agent runs three times per input. Table~\ref{tab:sopbench-k} gives five-fold cross-validation on the training split, averaged over three trace samples with one workflow generation each, from an earlier trace collection with the same protocol. Single procedure--method cells vary by as much as $50$ points across trace samples, so per-procedure gaps of a few points are within noise; we report intervals only for the macro test results (Table~\ref{tab:main}).

\begin{table}[!htbp]
\caption{\textbf{SOP-Bench test results by procedure.} Methods at $k{=}16$. The two repair columns start from $k{=}16$ \name\ workflows and from zero-shot workflows. No-reas.\ is \name with the reasoning removed. Bold marks the best result in each row.}
\label{tab:sopbench-sop}
\centering
\small \setlength{\tabcolsep}{2.5pt}
\adjustbox{max width=\linewidth}{%
\begin{tabular}{lrrrrrrrrrr}
\toprule
& & No- & Few- & Exec.- & Skill & Zero- & & & \name\ $+$ & Zero-shot \\
Procedure & \name & reas. & shot & only & Factory & shot & AWM & Agent & repair & $+$ repair \\
\midrule
customer\_service & 81.0 & 81.4 & 84.8 & 83.5 & 74.5 & 62.3 & 88.3 & \textbf{92.2} & 76.2 & 84.8 \\
dangerous\_goods & 75.5 & 79.4 & 78.7 & 78.7 & 78.7 & 78.7 & 78.7 & 73.8 & 91.2 & \textbf{95.6} \\
email\_intent & 83.5 & 83.2 & 85.7 & 85.3 & 74.6 & 67.7 & \textbf{100.0} & 98.2 & 95.7 & 93.5 \\
know\_your\_business & 51.1 & 60.0 & 59.3 & 60.0 & 37.8 & 26.7 & 53.3 & 60.0 & 67.4 & \textbf{82.2} \\
referral\_abuse\_v1 & \textbf{100.0} & \textbf{100.0} & \textbf{100.0} & \textbf{100.0} & \textbf{100.0} & \textbf{100.0} & \textbf{100.0} & \textbf{100.0} & \textbf{100.0} & \textbf{100.0} \\
referral\_abuse\_v2 & 94.9 & 94.9 & 99.3 & 90.5 & 33.3 & 82.7 & 98.6 & 96.9 & 99.7 & \textbf{100.0} \\
traffic\_spoofing & \textbf{80.8} & \textbf{80.8} & 74.7 & \textbf{80.8} & 56.9 & 71.7 & 76.1 & 80.1 & 77.8 & \textbf{80.8} \\
video\_classification & 78.4 & 78.4 & 86.6 & 68.0 & 28.5 & 51.5 & 81.1 & 82.8 & \textbf{94.8} & 89.0 \\
\midrule
Macro & 80.6 & 82.3 & 83.6 & 80.9 & 60.5 & 67.7 & 84.5 & 85.5 & 87.8 & \textbf{90.8} \\
\bottomrule
\end{tabular}}
\end{table}

\begin{table}[!htbp]
\caption{\textbf{SOP-Bench macro task success by number of traces}, five-fold cross-validation on the training split. Zero-shot uses no traces ($72.0\%$). Bold marks the best $k$ for each method.}
\label{tab:sopbench-k}
\centering
\small \setlength{\tabcolsep}{4pt}
\begin{tabular}{lrrrrr}
\toprule
Method & $k{=}1$ & $2$ & $4$ & $8$ & $16$ \\
\midrule
\name & 69.4 & 71.0 & 76.7 & 79.3 & \textbf{82.2} \\
\name, no reasoning & 72.7 & 74.7 & 75.4 & 79.4 & \textbf{81.4} \\
Few-shot & 73.5 & 75.4 & 77.0 & 79.5 & \textbf{82.9} \\
Execution-only & 64.9 & 66.6 & 67.0 & \textbf{75.8} & 75.7 \\
Skill Factory & 68.7 & 62.9 & 65.9 & 68.3 & \textbf{69.0} \\
AWM (runtime planner) & 81.2 & 82.7 & 82.7 & 84.0 & \textbf{84.4} \\
\bottomrule
\end{tabular}
\end{table}

\paragraph{Main comparison.} On the test inputs, \name leads few-shot only on traffic spoofing ($80.8\%$ against $74.7\%$), ties it on referral abuse v1, and trails it on the other six procedures (Table~\ref{tab:sopbench-sop}). Know-your-business ($51.1\%$ against $59.3$--$60.0\%$ for few-shot, execution-only, and \name without reasoning) and video classification ($78.4\%$ against $86.6\%$) account for most of the gap. In cross-validation, every method that uses traces scores higher at $k{=}16$ than at $k{=}1$ (Table~\ref{tab:sopbench-k}). Two procedures do not separate methods: referral abuse v1 is saturated (zero-shot scores $100\%$), and on dangerous goods almost every method without repair stays near $78.7\%$, because the SOP's own scoring rule disagrees with about a quarter of the labels. Know-your-business is hard even for the agent ($60.0\%$), yet a lookup on a single input field scores $73.3\%$ on the training inputs (leave-one-out), above every method's test score. In cross-validation at small $k$, \name's main failure is execution defects on traffic spoofing: loops that exhaust the step budget and calls to tools that the SOP names but the tool list lacks. Both disappear by $k{=}16$. AWM's lead comes mostly from email intent ($100\%$), where it reads each email at run time; the agent is close there ($98.2\%$) and leads on customer service ($92.2\%$).

\paragraph{Reasoning removal.} On the test inputs, \name without the reasoning scores $82.3\%$, $1.7$ points above full \name. It is ahead on know-your-business ($60.0\%$ against $51.1\%$) and dangerous goods ($79.4\%$ against $75.5\%$) and within $0.4$ points on the other procedures. In cross-validation, it is $3.3$ and $3.7$ points higher at $k{=}1$ and $2$, $1.3$ points lower at $k{=}4$, and within one point at $k{=}8$ and $16$ (Table~\ref{tab:sopbench-k}). SOP-Bench traces have $2$--$8$ tool-calling steps, and on five of the eight procedures every tool argument can be obtained by lookup from the task inputs and earlier results, so the reasoning has little to add that the actions do not already show.

\paragraph{Labels that contradict the SOP.} On customer service, the label is ``resolved'' exactly when latency after troubleshooting is at most $100$\,ms. The SOP additionally requires jitter of at most $30$\,ms and bandwidth at or above the subscribed plan, and every resolved case misses its plan's bandwidth slightly. In cross-validation at $k{=}16$, $12$ of the $15$ IRs contain the SOP's full guard; they note the bandwidth shortfall in the traces only as a caution, and the generated workflows escalate the resolved cases. With the bandwidth clause removed from the IR and everything else unchanged, \name rises from $82.7\%$ in a replicate run ($85.2\%$ in the main cross-validation run) to $92.8\%$ on the same trace samples ($29$ cells better, $5$ worse). Telling the workflow generator to prefer evidence over literal guards did not help ($80.6\%$). On email intent, the SOP allows an ``unable to decide'' answer that never occurs as a label; in cross-validation at $k{=}2$, \name gives it in $32$ of $279$ cells, and zero-shot never does.

\FloatBarrier

\section{Trace-producing agents on \tautwobench}
\label{app:source}

The agents that produce traces vary along two axes: the model (Claude Haiku 4.5 or Sonnet 4.6) and the reasoning format (native thinking or a two-pass ReAct loop). Figure~\ref{fig:model-iter}a shows the test success of the workflow built from each agent's traces. The agents' own success rates are measured over all their training runs, a different denominator, so we do not plot them next to the workflows. The main experiments use Sonnet 4.6 with native thinking, the only agent we also ran on the $40$ test inputs: there it scores $80.6\%$ with $12.1$ planning calls and $170$\,s per task (Table~\ref{tab:main}), above its $65.0\%$ success over the training runs.

Every agent spends $104$--$153$\,s and $14$--$30$ planning calls per task, which a compiled workflow removes (Table~\ref{tab:sources}). All values in that table are measured on the training runs.

\begin{table}[!htbp]
\caption{\textbf{Planning cost of each trace-producing agent}, per task on the \tautwobench training runs. Calls: planning calls; Time: planning time; native: native thinking; ReAct: two-pass ReAct.}
\label{tab:sources}
\centering
\small
\begin{tabular}{llrr}
\toprule
Model & Mode & Calls & Time (s) \\
\midrule
Haiku 4.5  & native & $17.2$ & $117$ \\
Haiku 4.5  & ReAct  & $30.3$ & $104$ \\
Sonnet 4.6 & native & $14.3$ & $153$ \\
Sonnet 4.6 & ReAct  & $30.0$ & $151$ \\
\bottomrule
\end{tabular}
\end{table}

\FloatBarrier
\section{Baseline implementations and fair comparison}
\label{app:baselines}

\paragraph{Choosing $k$.} Every method receives the same inputs where they apply: the SOP, the tool list, and the same pool of traces. Each method except AWM on \tautwobench chooses its number of traces $k$ by five-fold cross-validation on the training split before any test run; no method is tuned on the test split. On \tautwobench, cross-validation averages four workflow generations per fold, scores them with the same estimator as the test (task success among built workflows), and chooses among $k\in\{1,2,4\}$, breaking ties toward smaller $k$. We cap the candidates at $k{=}4$ because every trace costs a live agent run, so a method that needs few traces is the one worth deploying. Larger $k$, up to $32$, were swept afterwards as a sensitivity analysis, not as candidates (Appendix~\ref{app:raw-traces}). Choosing $k$ and the traces themselves without cross-validation is future work (Section~\ref{sec:limitations}). This selects $k^\star{=}4$ for \name, few-shot, execution-only, Skill Factory, and BPOP (whose ordering threshold is chosen the same way \citep{bpop}), and we also report AWM at $k{=}4$. \name's training scores are $44.3\%$ at $k{=}1$, $49.6\%$ at $k{=}2$, and $56.4\%$ at $k{=}4$, with build rates of $91.7\%$ at $k{=}2$ and $99.2\%$ at $k{=}4$. On SOP-Bench, cross-validation uses three trace samples per fold and compares $k\in\{1,2,4,8,16\}$ (Table~\ref{tab:sopbench-k}); $k{=}16$ is the best $k$, or within $0.1$ points of it, for every method, so Table~\ref{tab:main} reports every method at $k{=}16$.

\paragraph{Baseline implementations.} Zero-shot, few-shot, execution-only, and \name share one workflow-generation model and one generation prompt and differ in the evidence they receive and, for \name, in candidate selection and, on SOP-Bench, the input-field note (Appendix~\ref{app:dataset}). \name chooses among up to three candidates by re-executing them on its traces (Appendix~\ref{app:guarantees}); the other workflow-generation baselines keep their first valid workflow. In SOP-Bench cross-validation at $k{=}16$, giving the baselines the same selection raises few-shot from $82.9\%$ to $84.0\%$ and execution-only from $75.7\%$ to $81.2\%$, against $82.2\%$ for \name, so selection accounts for most ($5.5$ of $6.5$ points) of \name's margin over execution-only there. The external baselines follow their published methods on the same traces. We run Skill Factory \citep{skillfactory} as the planner-free code it produces, so that it can be compared directly with the other planner-free workflows. Its full system also makes one model call per task to choose a skill and fill its parameters, and falls back to a live agent when no skill applies (Table~\ref{tab:positioning}); we supply the parameters directly instead. AWM \citep{awm} keeps its runtime planner, and we report its runtime cost accordingly. BPOP learns only the order of tool calls and ends each run with a fixed message, so it cannot produce SOP-Bench's decision output, and we run it only on \tautwobench. Where the original papers left a setting unspecified, we chose the value that maximized the baseline's own cross-validation score.

\FloatBarrier
\section{What the workflow generator receives}
\label{app:authorinput}

\paragraph{Witnessed and unwitnessed rules.} The two shortened rules below illustrate the example in Section~\ref{sec:method-aggregate}. The first links a stated condition to a recorded action; the second records an alternative that the reasoning mentions but no supplied trace executes.

\begin{artifact}[unbreakable]{Two rules from the IR (shortened)}
Observed rule
  condition: check_network_status() reports "Airplane Mode: ON"
  action: toggle_airplane_mode()
  evidence: trace 1, step 2 (reason); step 3 (tool call)
Unwitnessed rule
  condition: check_network_status() reports "Mobile Data Enabled: No"
  action: toggle_data()
  reason: mobile data is required to send messages
\end{artifact}
Tool results provide the exact field names and text that the generated conditions must check. The unwitnessed rule shows what reasoning can add, but nothing guarantees that such a rule is correct.

The IR grows much more slowly than the full traces. From one trace to sixteen, it grows from $4.0$k to $10.4$k tokens, while the full traces themselves grow from $6.6$k to $87.3$k. It contains about $22$ decision rules up to $k{=}8$ and $31$ at $k{=}16$. The totals in Table~\ref{tab:carrier} include the SOP and tool definitions that every method receives.

\begin{table}[!htbp]
\caption{Workflow-generator input on \tautwobench, in mean tokens, for each number of traces $k$. ZS: SOP and tool list only, which every method receives; FS: $+\,k$ full traces; EO: $+\,k$ traces without reasoning; \name: $+$ the IR $\mathcal{I}$; $m$: decision rules in $\mathcal{I}$ (batch merge; incremental beyond $k{=}8$).}
  \label{tab:carrier}
  \centering
  \small
\begin{tabular}{lrrrrr}
    \toprule
    & \multicolumn{5}{c}{$k$} \\
    \cmidrule(lr){2-6}
    Input & 1 & 2 & 4 & 8 & 16 \\
    \midrule
    ZS    & \multicolumn{5}{c}{$14.1$k for every $k$} \\
    FS    & 20.7k & 24.9k & 35.6k & 56.5k & 101.4k \\
    EO    & 16.4k & 18.4k & 22.3k & 30.4k & 46.5k \\
    \name & 18.1k & 18.2k & 18.3k & 19.0k & 24.5k \\
    \midrule
    $m$   & 22 & 22 & 21 & 22 & 31 \\
    \bottomrule
  \end{tabular}
\end{table}

Given the same traces, the three methods give the workflow generator inputs of widely different sizes. We show all three for one \tautwobench telecom family, \texttt{mms\_issue} (MMS troubleshooting), built from the same two traces ($k{=}2$). The median generator prompt is $24{,}653$ tokens for few-shot and $18{,}356$ for execution-only, against the $14{,}120$ tokens that the SOP and tool list alone take (Table~\ref{tab:carrier}). This family's \name IR, shown below, is $4{,}083$ tokens, smaller than a single full trace, yet it states the branching structure that the generator would otherwise have to work out from the transcripts. The IR lists the opening steps, a repeated check (the completion check $\chi$), and its ordered decision rules, printed as numbered layers: \texttt{L1} to \texttt{L15} are the rules $\langle(g_1,\alpha_1),\dots,(g_m,\alpha_m)\rangle$ of Equation~\ref{eq:ir}. Each rule has a condition (\texttt{guard}), an action (\texttt{remedy}), and its supporting trace steps in $\langle\cdot\rangle$. A rule may also add a condition learned across traces (\texttt{unless}), an \texttt{UNWITNESSED} mark for a branch no trace reached, or an open question. The \texttt{PROBE} line describes a whole run, not a single step: the workflow re-applies Equation~\ref{eq:program} after every action, a remedy that clears its fault also clears its guard, and re-running the probe acts as the last rule, so repeated single-rule steps walk down the layers in order and apply every remedy whose guard still holds. We omit some layers and steps where marked and shorten long tool outputs.

\begin{artifact}{\name IR, \texttt{mms\_issue}, $k{=}2$ (4,083 tokens)}
ENTRY
  get_customer_by_phone(phone_number: str) -> customer_id, line_ids, status   <t0.0, t1.0>
  can_send_mms() -> initial MMS capability status                             <t0.1, t1.1>

PROBE  can_send_mms()
  fixed     "Your messaging app can send MMS messages."
  not fixed "Your messaging app cannot send MMS messages."
  On `not fixed`, descend all layers in order, applying every remedy whose guard
  holds; re-probe only after a full descent finds no further applicable layer.

L1 Airplane Mode ON
   why     Airplane Mode disables all radios; no cellular service means no MMS.  <t1.2>
   check   check_network_status()      guard "Airplane Mode: ON"
   remedy  toggle_airplane_mode()  ->  "Airplane Mode is now OFF."               <t1.3>

L2 SIM Card Missing
   why     A missing SIM means no cellular connection, a prerequisite for MMS.   <t0.2, t1.4>
   check   check_network_status()  (field "SIM Card Status")
   guard   "SIM Card Status: missing"
   remedy  reseat_sim_card()  ->  "SIM card re-seated successfully."             <t0.3, t1.4>
   !       The reseat response embeds the status bar inline; read the network
           type from it directly, no separate check_network_status call.         <t0.3, t1.4>

L4 Network Mode 2G (insufficient for MMS)
   why     2G cannot carry MMS; the phone must be on at least 3G.                <t1.4, t1.5>
   guard   status bar from a prior remedy shows network type "2G"
   unless  SIM seated and cellular service present (L2 remedy ran first)
   remedy  set_network_mode_preference(mode = "4g_5g_preferred")                 <t1.5>

L5 UNWITNESSED  Mobile Data disabled
   why     MMS requires mobile data; if the data switch is off, MMS cannot send.
   unless  Airplane Mode OFF (L1) and SIM seated (L2)
   guard   check_network_status() shows "Mobile Data Enabled: No"  (never triggered)
   remedy  toggle_data()
   ?       What does toggle_data() print on success?

... [layers L3, L6-L8 omitted] ...

L9 Account-side roaming not enabled on the matching line
   why     Abroad, MMS needs account-level roaming; without it data and MMS fail
           even when all device settings are correct.                    <t0.10, t0.11, t0.12>
   unless  Reached only when can_send_mms still fails after the full device
           descent (L1-L8).                                                      <t0.8>
   check   for line_id in customer.line_ids: get_details_by_id(line_id) until the
           result phone_number matches the user's number
   guard   matched line has roaming_enabled: false                              <t0.10>
   remedy  enable_roaming(customer_id, line_id) -> "Roaming enabled successfully" <t0.11>
   !       Account-side enable_roaming is distinct from device-side toggle_roaming
           and must be called first.                                     <t0.9, t0.10, t0.11>

... [layers L10-L15, all UNWITNESSED, omitted] ...

OPEN
  ?  When L1-L15 are exhausted and can_send_mms still fails, what non-escalation
     remedy should precede transfer_to_human_agents?              <t0.8, t0.12, t1.11>
\end{artifact}

\begin{artifact}{Few-shot prompt, same traces}
=== SHARED PREAMBLE (same for all three methods) ===
[SOP: tau2-telecom policy, ~22.8k chars, abbreviated]
  # Telecom Agent Policy ... technical support, overdue bills, line suspension,
  plan options ... [~22k chars of policy and troubleshooting trees]
[TOOL SCHEMAS: 43 tools] can_send_mms, check_apn_settings, check_network_status,
  get_customer_by_phone, get_details_by_id, reseat_sim_card, reset_apn_settings,
  toggle_wifi_calling, enable_roaming, ... (43 total)

=== DEMONSTRATIONS (the k=2 evidence: two raw traces, verbatim) ===
### DEMONSTRATION 1  (outcome: solved)
[step 0] THINK: The user is John Smith, phone number 555-123-2002, currently abroad
  in France, having trouble sending MMS messages. ... I must first identify the
  customer before providing technical support. [truncated]
  CALL get_customer_by_phone({"phone_number": "555-123-2002"})
  RESULT: {"customer_id":"C1001","full_name":"John Smith","account_status":"Active",
    "bill_ids":["B1001","B1002","B1003"], ...}
[step 1] THINK: The customer has been identified as John Smith (C1001). ... The customer
  is abroad in France, so I need to check their line details and also consider
  roaming. [truncated]
  CALL can_send_mms({})   RESULT: Your messaging app cannot send MMS messages.
[step 2] THINK: ... According to the SOP, I need to: 1. First ensure basic
  connectivity ... [truncated]
  CALL check_network_status({})
  RESULT: Airplane Mode: OFF | SIM Card Status: missing | Cellular Connection: no_service
    | Mobile Data Enabled: Yes | Data Roaming Enabled: Yes | ...
[step 3] THINK: ... SIM Card Status: **missing** ... [truncated]
  CALL reseat_sim_card({})
  RESULT: SIM card re-seated successfully. [status bar: Excellent | 5G | Data | 80%]
... [steps 4-13 omitted: check_apn_settings, check_wifi_calling_status,
    check_app_permissions, grant_app_permission, can_send_mms -> "cannot send",
    get_details_by_id x2 (matched line: roaming_enabled false), enable_roaming,
    can_send_mms -> "can send", FINAL_ANSWER] ...
### DEMONSTRATION 2  (outcome: solved)
... [second full trace, 13 steps, same THINK + CALL + RESULT format, stacked verbatim:
    get_customer_by_phone, can_send_mms, check_network_status (Airplane Mode: ON),
    toggle_airplane_mode, reseat_sim_card, set_network_mode_preference, ...,
    can_send_mms -> "can send", FINAL_ANSWER] ...

=== AUTHORING INSTRUCTION ===
Using the SOP, tool schemas, and demonstrations above, write one Python program
that reproduces the demonstrated policy for the mms_issue family.
\end{artifact}

\begin{artifact}{Execution-only prompt, same traces}
=== IDENTICAL PREAMBLE AND AUTHORING INSTRUCTION; every THINK block removed ===
### DEMONSTRATION 1  (outcome: solved)
[step 0] CALL get_customer_by_phone({"phone_number": "555-123-2002"})
  RESULT: {"customer_id":"C1001","full_name":"John Smith","account_status":"Active",
    "bill_ids":["B1001","B1002","B1003"], ...}
[step 1] CALL can_send_mms({})   RESULT: Your messaging app cannot send MMS messages.
[step 2] CALL check_network_status({})
  RESULT: Airplane Mode: OFF | SIM Card Status: missing | Mobile Data Enabled: Yes | ...
[step 3] CALL reseat_sim_card({})
  RESULT: SIM card re-seated successfully. [status bar: Excellent | 5G | Data | 80%]
... [steps 4-13 omitted; step 11 enable_roaming; step 12 can_send_mms -> "can send";
    step 13 FINAL_ANSWER] ...
### DEMONSTRATION 2  (outcome: solved)  ... [second trace, reasoning removed, elided] ...
\end{artifact}

\FloatBarrier
\section{Direct construction from full and execution-only traces}
\label{app:raw-traces}

This section gives the supporting experiments for Section~\ref{sec:results-full-traces}. \emph{Few-shot} gives the workflow generator complete traces, including the reasoning. \emph{Execution-only} removes the reasoning but keeps the tool calls, inputs, and results. \emph{Zero-shot} uses only the SOP and tool definitions. A construction builds when the generated code passes the checks needed to run it, which is different from solving the task.

\begin{figure}[!htbp]
\centering
\begin{subfigure}[t]{0.48\linewidth}
  \centering
  \includegraphics[width=\linewidth]{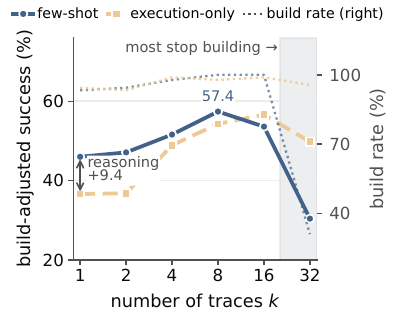}
\caption{\tautwobench}
  \label{fig:fewshot-tau2}
\end{subfigure}
\hfill
\begin{subfigure}[t]{0.48\linewidth}
  \centering
  \includegraphics[width=\linewidth]{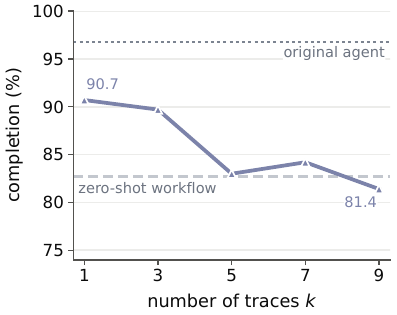}
\caption{\dset}
  \label{fig:fewshot-ent}
\end{subfigure}
\caption{\textbf{Adding full traces helps at first, then hurts the generated workflow.} (a) \tautwobench five-fold cross-validation on training inputs: build-adjusted success and build rate by number of traces. (b) \dset completion by number of traces, one workflow per point from a separate sweep, so its $k{=}3$ value differs from few-shot's $88.6\%$ in Table~\ref{tab:main}.}
\label{fig:fewshot}
\end{figure}

The simplest way to reuse an agent's past runs is to give their traces to the workflow generator as examples. We vary the number of traces $k$ on \tautwobench and on \dset and evaluate the resulting workflows. On \tautwobench we use five-fold cross-validation over the training inputs; on \dset we generate one workflow per procedure for each $k$ on a single split. Here we report \tautwobench results as \emph{build-adjusted} success, which counts a construction that fails to build as a failed task, so that a setting that stops producing runnable code cannot look like an improvement.

\paragraph{Full traces help only up to a point.} On \tautwobench (Figure~\ref{fig:fewshot}a), the generated workflow improves up to $k{=}8$ and declines afterwards; its build rate falls from $100\%$ at $k{=}16$ to $31\%$ at $k{=}32$. The decline coincides with longer inputs. Full traces grow the generator input from about $21$k tokens at $k{=}1$ to $101$k at $k{=}16$, and inputs that long reduce reasoning accuracy well before they reach the context limit \citep{inputlength,lostmiddle}. The pattern is stronger on \dset (Figure~\ref{fig:fewshot}b), where a single trace can exceed $20{,}000$ tokens while the SOP is a few thousand: completion falls from $90.7\%$ at $k{=}1$ to $81.4\%$ at $k{=}9$, below zero-shot's $82.7\%$ and far below the original agent's $96.8\%$.

\paragraph{Removing reasoning shortens the input but hurts at small $k$.} Dropping the reasoning $r_t$ and keeping only tool calls, inputs, and results roughly halves the input ($47$k tokens at $k{=}16$ against $101$k) and moves the peak from $k{=}8$ to $k{=}16$. Execution-only traces also keep building where full traces fail ($96\%$ against $31\%$ at $k{=}32$), but they need sixteen traces to reach their peak of $56.6\%$, while full traces reach $57.4\%$ with eight. The cost appears at small $k$: at $k{=}1$, full traces lead by $9.4$ points.

\section{Comparison of procedure construction and runtime control}
\label{app:positioning}

\begin{table*}[t]
\caption{\textbf{Procedure construction and runtime control.} Signal: evidence used for construction; Artifact: deployed artifact; LLM orch.: whether an LLM selects workflow steps or skills at run time, not counting models used inside fixed tools or nodes; Exec.\ cond.: whether the deployed artifact contains executable branch conditions, as opposed to conditions described in prompt text. ``Dispatch only'' means one model call to select a skill and fill its arguments. ``---'': not assessed. $^{\dagger}$TraceCompiler may still call an LLM to fill some arguments, but has no general runtime planner. $^{\star}$External baselines in our evaluation.}
\label{tab:positioning}
\centering
\small
\setlength{\tabcolsep}{4pt}
\adjustbox{max width=\textwidth}{%
\begin{tabular}{>{\raggedright\arraybackslash}p{0.20\textwidth}>{\raggedright\arraybackslash}p{0.25\textwidth}>{\raggedright\arraybackslash}p{0.27\textwidth}>{\raggedright\arraybackslash}p{0.10\textwidth}>{\raggedright\arraybackslash}p{0.08\textwidth}}
\toprule
Method & Signal & Artifact & LLM orch. & Exec.\ cond. \\
\midrule
AWM$^{\star}$ \citep{awm} & successful trajectories & prompt/context text & yes & no \\
ExpeL \citep{expel} & insights from successes and failures & prompt/context text & yes & --- \\
ReasoningBank \citep{reasoningbank} & successful and failed experiences & reasoning memory & yes & no \\
ASI \citep{asi} & verified success traces & skills (LLM-called) & yes & partial \\
SKILL-DISCO \citep{skilldisco} & success traces & skills (LLM-called) & yes & yes \\
NSI \citep{nsi} & execution traces & skills (LLM-called) & yes & yes \\
BPOP$^{\star}$ \citep{bpop} & action order & partial order & no & no \\
Skill Factory$^{\star}$ \citep{skillfactory} & verified solve traces & program (skill) & dispatch only & partial \\
TraceCompiler \citep{tracecompiler} & value provenance in tool traces & mostly deterministic workflow & residual$^{\dagger}$ & partial \\
Trace2Policy \citep{trace2policy} & VLM-reconstructed expert records and validation errors & human-refined deterministic Python (deployed variant) & no & yes \\
PLVR \citep{plvr} & teacher reasoning and action traces & program with model primitives and student & yes & yes \\
STAGE \citep{stage} & authoritative policy & executable graph with model-backed nodes & no & yes \\
\textbf{\name} & \textbf{SOP, reasoning $+$ execution} & \textbf{executable workflow} & \textbf{no} & \textbf{yes} \\
\bottomrule
\end{tabular}}
\end{table*}

Table~\ref{tab:positioning} expands the comparison in Section~\ref{sec:related}. It separates the evidence used for construction, the deployed artifact, and runtime control, because an executable artifact may still rely on a model to select skills or to carry out individual steps. Starred methods are the external baselines we evaluate; the others are compared from their papers only, without a controlled performance comparison against \name. Skill Factory is listed in its original deployment form; our evaluation supplies its skill parameters directly (Appendix~\ref{app:baselines}).

\section{One-time construction and repeated execution cost}
\label{app:workflow-cost}

Let $c_{\mathrm{plan}}$ be the original agent's cost of choosing steps in one run, $c_{\mathrm{tool}}$ the tool cost per run, and $C_{\mathrm{ind}}$ the one-time cost of building the workflow. If both methods have the same tool cost, running a procedure $n$ times costs
\begin{equation}
  \underbrace{n\,(c_{\mathrm{plan}}+c_{\mathrm{tool}})}_{\text{original agent}}
  \qquad\text{against}\qquad
  \underbrace{C_{\mathrm{ind}}+n\,c_{\mathrm{tool}}}_{\text{workflow } \pi}.
  \label{eq:cost}
\end{equation}
At comparable task quality, construction pays for itself after $n^\star=C_{\mathrm{ind}}/c_{\mathrm{plan}}$ runs. If the workflow changes the number or type of tool calls, the break-even should use the measured difference in total cost per run instead.

\paragraph{Example enterprise break-even.} We apply the \$2.32 construction cost estimated on \tautwobench to the enterprise setting. With the measured difference in total enterprise cost per task (\$0.031), about $75$ executions recover the construction cost, while the seven fixed workflows among the ten most frequently executed automations each ran $297$k--$528$k times over the three months covered by the platform log (Appendix~\ref{app:dataset}). The \$2.32 is an estimate from the public benchmark, not a measured enterprise cost, and enterprise construction uses Opus 4.8, which costs more; the actual break-even depends on the platform's construction costs.

\end{document}